\pdfoutput=1
\documentclass[a4paper,fleqn]{cas-sc}
\usepackage[authoryear]{natbib}
\usepackage{amsmath,amssymb,graphicx,booktabs,placeins}
\newcommand{\nodg}{\hphantom{^\dagger}}
\usepackage{url}
\usepackage{bookmark}
\usepackage{cmap}
\input{glyphtounicode}\pdfgentounicode=1

\begin{document}
\let\WriteBookmarks\relax
\title[mode=title]{Downside-Controlled Online Forecast Combination under Delayed and Revised Outcomes}
\shorttitle{Downside-Controlled Online Forecast Combination}
\author[1]{Minkyoung Kim}
\author[1]{Hyunjung Byun}
\author[1]{Yohan Lee}
\author[1]{Beakcheol Jang}
\cormark[1]
\affiliation[1]{organization={Graduate School of Information, Yonsei University}, city={Seoul}, postcode={03722}, country={South Korea}}
\cortext[1]{Corresponding author. E-mail address: bjang@yonsei.ac.kr}
\shortauthors{M. Kim et al.}
\begin{abstract}
Post-hoc correction adjusts a forecaster that cannot be retrained, such as a foundation model, but a correction fitted where errors are stable can hurt where they shift. We aim for downside control: not much worse than the starting forecast. We combine the frozen forecaster, a static corrector and an online corrector on the simplex, using only losses that mature after the horizon. Across seven benchmarks and four base models, two of them foundation models, the worst deterioration over 28 pairs at the main horizon is 0.15\% and gains reach 11.5\%. On day-ahead load for seven European bidding zones it lowers mean MSE in all seven zones, while single correctors raise mean MSE by up to 102\% where the published forecast is most accurate. Three empirical conditions on expert speed, stream length and outcome alignment, each fixed by a documented failure, delimit its scope. Learning from the provisional outcome improves four zones on the settled one; learning on the settled outcome restores all seven.
\end{abstract}

\maketitle
\hypersetup{pdfauthor={Minkyoung Kim, Hyunjung Byun, Yohan Lee, Beakcheol Jang}, pdfsubject={}}
\section{Introduction}
\label{sec:intro}

Pretrained foundation models such as Chronos \citep{chronos2024} and TimesFM \citep{das2024timesfm} are served behind interfaces that expose forecasts but not weights. Forecasters validated inside regulated or safety-critical pipelines are expensive to revalidate after any change. In both cases, the practical question is how to improve a model that cannot be retrained, not how to train a better one. We call such a forecaster frozen.

Post-hoc residual correction leaves the forecaster frozen and trains a small module that adjusts the forecast \citep{Kim2022Rescal,deltaadapter2026,Liu2025PIR}. A correction layer is orders of magnitude smaller than the base model and can be attached or detached without touching it. Its risk is less often discussed. A corrector fitted where the error structure is stable can be actively harmful where it is not. In our experiments with a DLinear base model, a corrector fitted on held-out data improves ETTm2 by 7.5\% but deteriorates Weather by 5.2\%. A published test-time adaptation method \citep{tafas2025} improves Exchange by 17.0\% but deteriorates ETTm2 by 7.3\% on the same base model. For a frozen forecaster that is already in production, an intervention that sometimes makes things worse is difficult to justify regardless of its average gain.

Forecast combination limits the loss when one of the combined forecasts fails: weights formed from past errors go back to \citet{bates1969}, and restricting them to the simplex, so that they are non-negative and sum to one, helps when the experts are highly correlated \citep{radchenko2023similar}. Online weighting extends this to streams, with the loss of one period observed before the weights of the next are formed \citep{devaine2013experts,berrisch2024crps}. Deploying a frozen forecaster departs from that setting in two ways. First, the outcome of a multi-step forecast arrives only after the horizon has elapsed, so every weight update is delayed. Second, in operational data the outcome itself is revised: a provisional outcome is published quickly and a settled one months later, the real-time data problem of macroeconomic forecasting \citep{croushore2001realtime}, and the situation that intercept correction was designed for \citep{clements1996intercept,castle2024shifts}. The load study below measures whether a combination whose experts learn from the provisional outcome keeps its gains when judged on the settled one.

Our design objective is downside control rather than peak accuracy: on any series, the corrected forecast should not be much worse than the forecast it starts from. We first test whether the residuals of frozen base models contain structure that a corrector can reach, and which descriptors of the data recover it. On five standard benchmarks, the residuals are autocorrelated on every dataset and every run, strongly so only on Weather. Yet a ridge regression that predicts them from per-sample statistics of the input window has negative cross-validated $R^2$ everywhere. Next, a pre-specified test examines the interface that feature-conditioned and language-model-conditioned correctors share \citep{ctrl2026}: a low-dimensional conditioning vector that modulates the corrector, using per-sample oracle probing. At the specified probing budget, the ceiling this interface can reach stays under 5\% on all five datasets. The residuals are serially dependent, but the window summaries we test do not capture that dependence.

To reach that dependence, the corrector reads the error sequence itself and uses each realized error once it matures. Any single corrector rests on an assumption that can fail, so its weight should come from its realized performance rather than be fixed in advance. This makes the problem one of forecast combination. The experts are the frozen base model, a static trust-region corrector and an online corrector. At every origin, the gate's weights decide how strongly the layer changes the frozen forecast.

The two correctors solve different problems. The static corrector learns a stable residual pattern from abundant training data, but it cannot react when that pattern changes. The online corrector tracks recent drift from scarce matured outcomes, but it can overreact. On the trained base models, the static corrector alone already meets the objective. Where the base model is weaker, as with frozen statistical forecasters and with foundation models on the ETTm series, the static corrector gains only a fraction of what the online corrector reaches. Instead of choosing between them, a gate sets the weights of all three experts from matured losses. Each update multiplies every weight by a factor that shrinks with that expert's loss and renormalizes to the simplex. The gate is warm-started by replaying the same update on held-out data.

The gate inherits the regret bound of the exponentially weighted average forecaster \citep{cesabianchi2006}. That bound limits the gap between the combination's accumulated loss and the loss of the best single expert in hindsight. On every stream admitted under the conditions of Section~\ref{sec:conditions}, the measured downside of the layer stayed small relative to the forecaster it starts from. Where the base model is strong, the gate keeps most of its weight on it and the corrections are small. Where the base model is weak, the gate can move nearly all of its weight to a corrector, so that the layer replaces the forecaster rather than adjusting it. The regret bound is asymptotic, and the worst-case figures reported below are measured outcomes of runs under those conditions, not consequences of the bound.

We evaluate on seven multivariate benchmarks with four frozen base models, two trained (DLinear, PatchTST) and two zero-shot foundation models (Chronos-Bolt, TimesFM). Across 28 dataset-model pairs, the worst deterioration is 0.15\% and gains reach 11.5\%, and the same downside control holds on foundation models the correctors were never designed against. We then apply the same layer to a forecaster that nobody outside its issuer can retrain: the day-ahead load forecast that European transmission system operators (TSOs) publish for seven bidding zones. The combination improves every zone on the outcome it learns from. Where the forecast is already accurate, the held-out and online correctors alone cost 102 and 85\%. The static corrector, which deteriorates on no benchmark pair, deteriorates on three zones. The gain ranges from near zero to 57\% and is broadly larger where the forecast is less accurate.

For intervals, we wrap the combined forecast in adaptive conformal calibration \citep{gibbs2021}, which tunes the interval width online from the realized coverage. By the Winkler score, which penalizes an interval for each outcome falling outside it, the adaptive intervals beat a single split-calibrated width on 13 of the 14 trained-base-model pairs and in every load zone, at the price of wider intervals on the benchmarks.

The applicability conditions come from experiments that failed. First, experts must be quasi-static on the timescale of the maturation delay. Second, the stream must be long enough to host the warm-start layout and to benefit from it. Third, the outcome the experts learn from must be the outcome they are judged on. The two load outcome versions test the third condition. Experts that learn from the provisional outcome and are judged on the settled one share a bias that no combination confined to the simplex can remove. Learning on the settled outcome at delays of zero and thirty days restores the improvement in every zone.

We make the following contributions.
\begin{itemize}
\item We audit the residuals of frozen base models. With oracle probing on a DLinear base model, we measure how much a corrector can gain through the conditioning interface that feature-conditioned and language-model-conditioned correctors share. At the pre-specified budget, that interface leaves little headroom on those series.
\item We develop an online combination of a frozen forecaster with a static and an online corrector, designed for downside control. It learns from matured outcomes only, and its weights are warm-started on a disjoint slice of the held-out data that covers the opening of the stream, before a delayed gate has scored anything. We verify it on benchmarks and on operational load forecasts.
\item On real load data, we compare learning from a provisional outcome with learning from a settled one. We state three applicability conditions, on expert speed, stream length and outcome alignment, each fixed by a failure test on the evaluated streams.
\end{itemize}

\section{Related work}
\label{sec:related}

\subsection{Forecast combination and expert aggregation}
\label{sec:rw-combination}

Combining forecasts of one target with weights formed from their past errors goes back to \citet{bates1969}, and unconstrained regression weights to \citet{granger1984improved}. The surveys of \citet{clemen1989}, \citet{timmermann2006combinations} and \citet{wang2023combinations} record how consistently simple combinations improve on their experts. Which scheme to trust in a given deployment is less settled: \citet{wang2023combinations} find no consensus on which combination method performs best in a specific setting, and \citet{timmermann2006combinations} traces unstable weights to the nonstationarity that motivates combining in the first place. Sections~\ref{sec:opsd} and~\ref{sec:conditions} return to that question for streams with delayed and revised outcomes. The advantage is not automatic. \citet{koning2005m3} reexamined the M3 competition and found that its one combination beat two of its three experts only a little more than half the time. They judged the competition's conclusion on combination unproven.

This literature shapes the gate. Weights estimated from a finite error history are themselves uncertain, and that uncertainty adds a term $\mathrm{var}(\hat w)\,\mathrm{var}(y_1 - y_2)$ to the variance of the combined forecast. Here $\hat w$ is the estimated weight and $y_1 - y_2$ the difference between the two experts' forecasts. This is why estimated optimal weights are so often worse than an equal average \citep{claeskens2016puzzle}. \citet{radchenko2023similar} show where that term is large. Suppose the experts are highly correlated, with correlation $\rho$, and have similar error standard deviations $\sigma_1$ and $\sigma_2$. Then the optimal weight $w^*$ falls outside $[0,1]$ and becomes negative once $\rho > \sigma_2/\sigma_1$. The variance of its estimate grows as $1/(1-\rho^2)$. Forcing the weights back into $[0,1]$ therefore trades a small bias for a large reduction in variance. The gate never estimates $w^*$. It sets the weights by multiplicative updates on realized losses. The weights stay non-negative and sum to one, so the trimming is built in. Its worst-case behavior rests on the regret bound rather than on the accuracy of a covariance estimate. Section~\ref{sec:main-results} checks this regime on the two correcting experts.

Time-varying and online weighting has since been developed for point and probabilistic forecasts. \citet{berrisch2024crps} and \citet{lee2026breaks} form the weights from realized errors. \citet{vandermeer2024crps} learn a nonlinear pool of predictive distributions online from gradients of the continuous ranked probability score, approaching the optimal combination in hindsight on stationary synthetic series and outperforming it on nonstationary wind power data. \citet{gibbs2024conditional} form conditionally optimal weights from information available at the forecast origin. \citet{bernaciak2024discounting} score experts by discounted past loss. In their multilevel scheme, the effective discount rate, which sets how fast the weights move, itself varies over time. \citet{devaine2013experts} analyse the specialist aggregation rule of \citet{freund1997specialists} and the fixed-share rules of \citet{herbster1998}. They apply them to one-day-ahead electricity load, with experts that abstain outside their regime, which makes their study the closest precedent for the load study of Section~\ref{sec:opsd}. \citet{hassoun2026kairosis} aggregate forecasts of an outcome that is resolved only at the end of the period. No loss is observed while the weights are formed, so their weighting follows change points in the forecast stream instead. The gate combines both features: realized losses drive its weights, but the loss of the forecast issued at origin $o$ becomes observable only $H$ periods later. The warm start of Section~\ref{sec:gate} and the quasi-static condition of Section~\ref{sec:quasistatic} are the two consequences of that delay. Repeated updating also has a cost on the forecast side. \citet{vanbelle2023stability} define rolling origin forecast instability, the variability in forecasts for one target period as the origin advances. They train against a composite loss that penalizes it alongside accuracy. \citet{caljon2026dynamic} improve that trade-off by weighting the two loss components dynamically during training. \citet{godahewa2025stability} separate vertical from horizontal stability and obtain both, for any base model, by linear interpolation between forecasts from adjacent origins and adjacent horizons. That line of work stabilizes the forecasts a model issues as data accrue. Here the base forecasts are frozen, and the outcomes arrive late and revised.

Learning from one release of a series and being judged on a later one is the real-time data problem of macroeconomic forecasting. \citet{croushore2001realtime} built the vintage archive that made it studiable, \citet{croushore2003vintage} found published conclusions reversed on other vintages, and \citet{croushore2011frontiers} surveys the decade that followed. \citet{koenig2003realtime} are closest to the protocol here. They argue against fitting on the latest vintage, because it gives the estimator information that no forecaster held at the origin. On the load data of Section~\ref{sec:opsd}, the issuer publishes the two outcome versions side by side, so they can be crossed directly.

The gate itself is the exponentially weighted average forecaster from the literature on prediction with expert advice. There, a learner combines several predictors and is judged against the best of them in hindsight, and the regret bound is standard \citep{freund1997,cesabianchi2006}. Asymmetric variants bound the loss against a designated comparator instead: a larger prior weight on one expert tightens the bound against that expert \citep{cesabianchi2006}, and mixing toward a fixed benchmark yields constant regret to the benchmark while remaining competitive with the best expert \citep{evendar2008regret,sani2014easy}. The gate turns a set of correctors, none of them reliable everywhere, into a layer with a controlled downside. Its interaction with delayed outcomes requires care. Adaptive conformal inference \citep{gibbs2021} widens or narrows the interval after each origin according to whether the last outcome fell inside it. This delivers the nominal coverage in the long run even when the errors are not exchangeable. Conformal PID control \citep{angelopoulos2023} pursues the same aim by treating interval production as a control problem. We use the adaptive tracker as the interval layer over the combined forecast, and claim only long-run coverage.

\subsection{Correcting a frozen forecaster}
\label{sec:rw-correction}

Adjusting a frozen model's forecasts by its recent realized errors is intercept correction in econometric forecasting \citep{clements1996intercept}. \citet{castle2015robust} list it among the established responses to a location shift, a change in the level the series reverts to. After such a shift, a model estimated on the earlier level keeps forecasting toward it. An adjustment formed from the latest errors removes much of the resulting bias, at the price of added variance when no shift has occurred. \citet{castle2024shifts} show that a model describing a shift well in sample can forecast worse than one that ignores it. Their remedy is an added term that lets the data decide which kind of shift occurred. The online corrector of Section~\ref{sec:experts} is a multivariate, horizon-wise form of this error-driven adjustment, and the static corrector is its train-split analogue. The gate plays the role of the deciding term. It weights the adjusted and unadjusted forecasts by realized losses instead of committing to either, so the mechanism that gives weight to a helpful correction also withdraws weight from a harmful one.

ResCAL \citep{Kim2022Rescal} estimates the residuals of a frozen traffic forecaster from previous errors and shows that errors commonly attributed to noise are partly predictable. Recent work extends the idea to general benchmarks. \citet{deltaadapter2026} learn small modules on the input and the output of a frozen model and bound how far they may move the forecast. They also add calibrators that adjust the quantiles of the predictive distribution. \citet{Liu2025PIR} identify the least reliable instances and revise them from covariates and from similar instances retrieved from the historical record. \citet{Chen2024Calibration} attach a calibration scheme to a trained forecaster in the same spirit. \citet{Liang2024Minusformer} build residual subtraction into the architecture itself, a within-training counterpart of post-hoc correction. Once fitted, these correctors are fixed functions. Where one varies at test time, it varies with the input, not with the errors the forecaster is currently making. Even the online variant of \citet{deltaadapter2026}, which keeps updating the adapter as outcomes arrive, fixes in advance how far the correction may move the forecast, instead of setting it from the correction's realized performance. The static expert here belongs to the fitted-once family, and the gate supplies the missing feedback.

A second family updates at test time. TAFAS \citep{tafas2025} adapts a forecaster during deployment from partially observed outcomes through gated calibration modules. PETSA \citep{petsa2025} calibrates inputs and outputs through low-rank adapters with a frequency-aware loss. Both leave the forecaster frozen and update add-on parameters, as we do. They differ from the setting here in two ways. The first is the size of the updated surface: 50.4 million parameters for the calibration modules of \citet{tafas2025} on Electricity, against 37,252 here. The second is the outcome protocol. Both consume partially observed outcomes, and PETSA also uses delayed complete ones, while the layer here uses fully matured outcomes only. Gradient-based test-time adaptation from the vision literature \citep{wang2021tent} updates parameters by entropy minimization under an evaluation protocol different from standard forecasting benchmarks. Neither TAFAS nor PETSA provides prediction intervals.

ORCA \citep{dai2026orca}, concurrent work on black-box adaptation of foundation forecasters, learns a residual adapter from the input window and the base forecast. It weights the adapted and the base forecasts by a softmax over their exponentially smoothed errors, under the same matured-outcome rule used here. The two designs share the outcome protocol and the two-expert safety objective. The study here also covers the warm start of the weights, the heterogeneous expert library, the interval layer, and the ceiling measurement of Section~\ref{sec:audit}. The two papers report worst-case figures under different normalizations, a per-channel z-scale here and a dataset-level scalar there, so the figures are not directly comparable.

\subsection{Base models, normalization and conditioning}
\label{sec:rw-base}

The base models a correction layer wraps span three families. Linear and decomposition models such as DLinear \citep{Zeng2023DLinear} and patch transformers such as PatchTST \citep{Nie2023PatchTST} are the trained base models used here. Pretrained forecasters \citep{chronos2024,das2024timesfm}, surveyed by \citet{Liang2024Foundation}, supply zero-shot forecasts without dataset-specific fitting. The method does not depend on this choice, and we evaluate it on base models from each family.

Around these base models, normalization layers such as RevIN \citep{Kim2022ReVIN} rescale the input and the output so that a drifting mean and variance do not reach the model. Online ensembling under concept drift \citep{Wen2023OneNet} adapts the forecaster itself. These methods reduce the effect of shift but do not remove the residual that remains, and a corrector acts on that residual. Non-stationarity is also why a fixed corrector is not reliable everywhere.

A separate line of work positions language models as components of forecasting pipelines, either as numerical predictors \citep{Jin2024TimeLLM,zhou2023onefitsall}, as cross-modal aligners \citep{Liu2025CALF}, or as reasoners that emit control signals for a downstream module \citep{ctrl2026}. Systematic evaluations report that removing or replacing such components rarely worsens accuracy \citep{Tan2024LLMUseful} and that language-model forecasters are fragile to small perturbations \citep{park2025revisiting}. The ceiling test of Section~\ref{sec:audit} offers a mechanism for these observations in the correction setting. At the audit's budget, the conditioning route that those designs share has almost no reachable headroom, so differences between controllers reveal little about the controllers themselves.

\section{Residual audit and ceiling test}
\label{sec:audit}

Post-hoc correction of a frozen forecaster is only useful if the forecaster's errors contain structure that a corrector can reach. We ask two questions before building any method. First, do the errors of frozen base models on standard benchmarks carry any predictable structure? Second, if they do, can a corrector reach that structure through the route several published designs use? In that route, a short conditioning vector rescales and shifts the corrector's internal quantities through a conditioning interface \citep{ctrl2026}. In CTRL, one eight-dimensional vector serves the whole dataset and is adjusted only when a shift is detected. Sections~S2 and~S3 report the audit and the ceiling test in full.

The audit examines the residuals of a frozen DLinear base model \citep{Zeng2023DLinear} on ETTh1, ETTh2, ETTm1, ETTm2 and Weather with three diagnostics summarized over channels and five runs (Table~S5). The residuals are taken on the fit region of the held-out split, the region that Section~\ref{sec:setup} reserves for fitting correctors and that is disjoint from the test stream. A Ljung--Box test \citep{ljungbox1978} asks whether the one-step-ahead residuals are white noise. A ridge regression asks how much of the horizon-mean residual can be predicted from summary statistics of the input window. The statistics come from a moving-average seasonal-trend decomposition, and the regression is scored by cross-validated $R^2$. The lag-one autocorrelation reports how much of a residual carries over to the next step. The first and third diagnostics target the error sequence in time; the second targets what a single window says about its own error. The Ljung--Box test rejects whiteness on every dataset and on every run; the largest $p$ across channels and runs at lag 24 is $8.2\times 10^{-3}$. The errors are therefore serially dependent. The ridge $R^2$, however, is negative on all five datasets, between $-0.149$ and $-0.815$ at the channel median. The regression thus predicts the residual worse than its own training mean does. The summary statistics we test, which are what a conditioning vector would carry, recover nothing usable about the horizon-mean residual. The dependence therefore lies in the time axis of the error sequence, not in the window summaries we test.

\begin{figure}[pos=!ht]
\centering
\includegraphics[width=0.8\columnwidth]{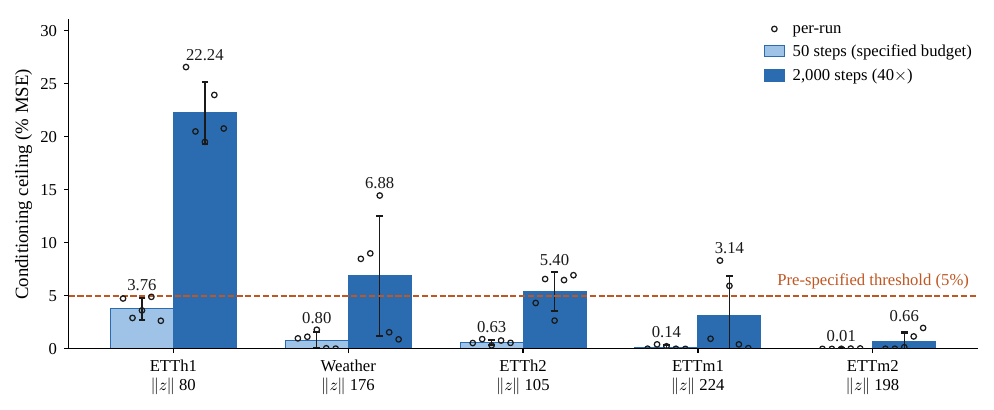}
\caption{Conditioning ceiling against the number of oracle optimization steps per sample. Bars are five-run means with one standard deviation; open markers are the five individual runs.}
\label{fig:oracle}
\end{figure}

The audit is correlational, so we test the conditioning route directly with a ceiling test. The decision rule was fixed before any run: if conditioning cannot improve a well-designed corrector by more than 5\% of mean squared error (MSE) on at least two of five datasets, we abandon it. The corrector passes each decomposition component of the forecast through its own linear map over the horizon, $\mathrm{Linear}(H,H)$, and a conditioning vector $z \in \mathbb{R}^{8}$ rescales and shifts the intermediate quantities of that map \citep{Perez2018FiLM}; it is the online corrector of Section~\ref{sec:method} with this interface added. The corrector is fitted on the first two thirds of the fit region, with its early-stopping tail inside that part, and probed on the last third, which it never saw. Any learned controller would have to produce $z$ from the input window. Instead of reporting what one such controller achieves, we ask how much the interface could give at best. For each probe sample, we set $z$ by optimizing it against that sample's own outcome, which no deployable method could do. We call this per-sample oracle probing. The resulting error is what an outcome-informed oracle reaches through this route at a stated optimization budget, whether it sets one vector per sample or one for the whole dataset. We call that value the ceiling. It is reached by optimization rather than proved optimal, so it can rise with the budget (Table~S6 and Fig.~\ref{fig:oracle}).

Under the pre-specified budget of 50 optimization steps per sample, the ceiling does not exceed 5\% on any of the five datasets; the largest is 3.76\% on ETTh1. At 2{,}000 steps, 40 times that budget, the mean ceiling exceeds 5\% on three datasets, ETTh1 at 22.24\%, Weather at 6.88\% and ETTh2 at 5.40\%, and on the last two the standard deviation across runs is larger than the margin over the threshold. Those values are reached only after the oracle drives $\lVert z \rVert$ to between 80 and 224, far outside the scale the pre-specified budget reaches, and the oracle sets $z$ against the outcome it is scored on. The audit shows that the window summaries a controller could read do not predict the residual, so the headroom at the extended budget is not one a controller acting before the outcome can claim.

Together the two tests locate why conditioned correction underperforms here. The dependence the audit finds can be tracked over time, but it cannot be read off a summary of a single input window. At the pre-specified budget the conditioning route offers less than 5\% on every dataset, so this paper does not take it; the input window itself is not in question. Correctors that read the window directly, as in \citet{dai2026orca} and \citet{Liu2025PIR}, are unaffected. CTRL \citep{ctrl2026} reports that a language-model controller produces better control signals than a controller that maps decomposition features to the same signals. A low ceiling on the conditioning interface is a separate statement: it is what either controller reaches through the route they share at that budget. The two hold together, and the second is consistent with broader evidence that language-model components add little to numerical forecasting pipelines \citep{Tan2024LLMUseful,park2025revisiting}. These results call for a corrector that reads the error sequence itself as it becomes observable, and for a mechanism that limits the loss where the corrector is wrong. Section~\ref{sec:method} builds both.

\section{Method}
\label{sec:method}

Section~\ref{sec:audit} calls for a corrector that reads the realized error sequence as it becomes observable. Any such corrector can do harm where its assumptions fail. We therefore build the method from three parts: a delayed-outcome streaming protocol that defines what is observable and when, a small library of complementary correctors, and a multiplicative-weights gate that allocates weight among them from realized losses only. Throughout, $E_0$ is the frozen base model, never fitted here; $E_1$ is the static corrector, fitted once on the training split and then frozen; $E_2$ is the online corrector, initialized on held-out data and updated as outcomes mature. The gate with $K{=}2$ combines $E_0$ and $E_2$; the gate with $K{=}3$ combines all three.

\begin{figure}[pos=!ht]
\centering
\includegraphics[width=0.9\columnwidth]{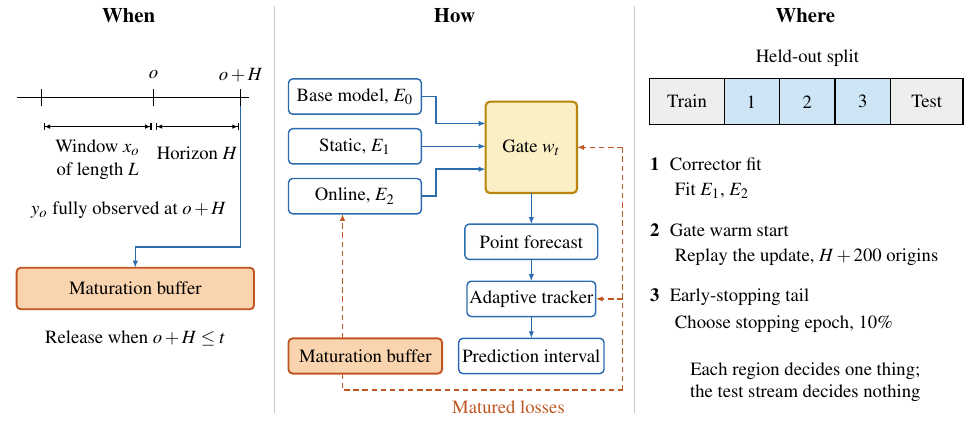}
\caption{The layer under delayed feedback. When: a forecast issued at origin $o$ has its outcome fully observed only at $o+H$, and the maturation buffer releases that origin once $o+H$ has passed. How: the gate weights the frozen base model and the two correctors, and every adaptive component reads matured losses through the buffer and through no other path (dashed). Where: the held-out split is divided into three disjoint regions, each deciding one thing.}
\label{fig:overview}
\end{figure}

\subsection{Setting and maturation protocol}
\label{sec:protocol}

A frozen base model $f_\theta$ maps an input window $x_o$ of the last $L$ observations ending at origin $o$ to a forecast over a horizon of $H$ steps and $C$ channels, one channel per series of the multivariate target,
\begin{equation}
\hat{y}_o = f_\theta(x_o) \in \mathbb{R}^{H \times C}.
\label{eq:basemodel}
\end{equation}
We evaluate on a rolling origin: the origin advances one step at a time through the test period, and a forecast is issued and scored at each position. The outcome $y_o$ of origin $o$ spans wall-clock steps $o{+}1$ through $o{+}H$, so it is fully observed only once the stream reaches origin $o{+}H$. A maturation buffer releases origin $o$ to every adaptive component at the current origin $t$ exactly when
\begin{equation}
o + H \leq t.
\label{eq:maturation}
\end{equation}
No component of the layer reads test outcomes through any other path, and the release condition is checked at runtime. On release, the gate scores the expert forecasts that were issued and stored at origin $o$ and updates its weights from those losses. The online corrector also updates from the same matured outcome, but this update changes only the forecasts it issues afterwards, never a loss already scored.

Partially observed horizons are never used, so the protocol is more conservative than the partial-outcome schemes of \citet{tafas2025} and \citet{petsa2025}. Section~\ref{sec:experiments} quantifies the loss and the gain from that choice. The same release rule is used by the online variant of \citet{deltaadapter2026}. \citet{actnow2024} classify any use of future signals to update a model as information leakage. \citet{lau2025} show that when the origin advances one step at a time, the window being scored overlaps steps already used for parameter updates, and this flatters the reported accuracy. For throughput, we process the stream in chronological chunks. Chunking only delays updates further, and it changes headline MSE by less than 0.05\% with no consistent sign, so per-step and chunked schedules are interchangeable in practice.

\subsection{Expert library}
\label{sec:experts}

Three experts produce candidate predictions at every origin. $E_0$ is the base model itself, the expert that leaves the forecast unchanged. $E_1$ adds a static trust-region corrector in the style of \citet{deltaadapter2026},
\begin{equation}
y^{(1)}_o = \hat{y}_o + \delta\, A(\hat{y}_o, x_o), \qquad \lVert A \rVert_\infty \leq 1.
\label{eq:static}
\end{equation}
Here $A$ is a two-layer network whose tanh output lies in $[-1,1]$ elementwise. It is fitted once on the training split against the frozen base model's in-sample residuals and then frozen. The scalar $\delta$ caps how far the correction may move the forecast, and its value is given in Section~S1. $E_2$ adds an online sequence corrector with the per-component $\mathrm{Linear}(H,H)$ architecture of Section~S3. It is initialized on the held-out split and then updated by one gradient step on each batch of matured outcomes.

The two correctors differ in the residuals they learn from. $E_1$ is fitted to the base model's in-sample residuals over the whole training split. These residuals are many, but they all come from the data the base model itself was fitted on, and $E_1$ cannot change after fitting. $E_2$ learns from the far smaller set of matured out-of-sample errors that the stream reveals, so it can follow drift. In the experiments, $E_1$ is better on datasets whose error structure is static and $E_2$ on datasets whose error structure shifts. The gate reduces the need to choose between them in advance. Freezing $E_1$ and updating $E_2$ only at a slow cadence also keeps both experts nearly fixed on the timescale of the maturation delay.

\subsection{Hedge gate with held-out warm start}
\label{sec:gate}

Fig.~\ref{fig:overview} shows the resulting system. The gate produces no forecast of its own. It decides how much weight to give the forecasts that already exist. It raises the weight of the expert whose matured losses have been smaller and lowers the others. The prediction is the convex combination
\begin{equation}
\tilde{y}_t = \sum_{k=0}^{K-1} w_{k,t}\, y^{(k)}_t, \qquad w_t \in \Delta^{K-1},
\label{eq:combination}
\end{equation}
where $\Delta^{K-1}$ is the set of weight vectors that are non-negative and sum to one. The weights are maintained by the multiplicative-weights update of the exponentially weighted average forecaster \citep{cesabianchi2006}, the Hedge update,
\begin{equation}
w_{k,t+1} \propto w_{k,t} \exp(-\eta\, \ell_{k,t}), \qquad \ell_{k,t} = \frac{1}{\sigma^2_{\mathrm{ho}}} \cdot \frac{1}{|\mathcal{M}_t|} \sum_{o \in \mathcal{M}_t} \lVert y_o - y^{(k)}_o \rVert^2_2 / (HC).
\label{eq:hedge}
\end{equation}
Here $\mathcal{M}_t$ is the set of origins matured at step $t$ under (\ref{eq:maturation}). The constant $\sigma^2_{\mathrm{ho}}$ is the base model's MSE on the held-out split, fixed before streaming. Dividing by it puts the losses of every dataset on the same scale, so one learning rate $\eta$ serves all of them. We use $\eta = 0.1$ throughout, and varying it over a sixfold range moves MSE by at most 1.14\% (Section~\ref{sec:ablations}).

This is the exponentially weighted average forecaster of \citet{cesabianchi2006}, whose regret against the best single expert over a stream of $T$ origins satisfies
\begin{equation}
\sum_{t=1}^{T} \ell_{\tilde{y},t} - \min_{k} \sum_{t=1}^{T} \ell_{k,t} \le c\,\sqrt{T \log K}.
\label{eq:regret}
\end{equation}
Here $c$ depends on the loss range. The bound holds for losses confined to a bounded range and a learning rate chosen for the horizon. It is asymptotic and concerns regret: it limits how much loss the combination can accumulate relative to its best expert in hindsight, and it sets no floor for any single stream. Because the base model is itself an expert, (\ref{eq:regret}) also bounds the combination's accumulated loss relative to the frozen base model under the same conditions, and this cumulative comparison is the only protection the theory gives. The worst-case figures reported below are measured outcomes, not consequences of (\ref{eq:regret}). In the experiments, $\eta$ is fixed at one value across datasets and horizons rather than tuned to the test stream, so the runs do not use the learning rate that would optimize the finite-horizon constant. Panel C of Table~S13 gives the sensitivity. The comparator in that statement is the base model alone, a fixed expert by construction. The analysis behind (\ref{eq:regret}) places no condition on the other experts, whose losses enter only as observed sequences. An expert that keeps training as outcomes mature therefore does not weaken the bound against the base model; what it can do is make the scoring stale, which Section~\ref{sec:quasistatic} shows.

The classical setting behind (\ref{eq:regret}) assumes that each expert's loss is observed as soon as it is incurred and that losses lie in a known range. Our setting departs on both counts. Losses become observable only at maturation, so every update acts on losses at least $H$ steps old. Under a fixed delay, the same update keeps a bound of the form (\ref{eq:regret}), with the stream length inflated by a factor of the order of the delay \citep{weinberger2002delayed,joulani2013delayed}. This is why the length of the stream relative to $H$ matters. The constant in (\ref{eq:regret}) also depends on the range of the losses, and the normalized loss in (\ref{eq:hedge}) is limited only by the largest error the stream produces, which is not known in advance. The delay does break a separate assumption: that an expert's loss when it is scored equals its loss when its forecast is used. Section~\ref{sec:quasistatic} shows an expert that changes over the delay window and is therefore weighted on a version of itself that no longer exists. Section~\ref{sec:conditions} states the operating condition that rules this out. The worst-case figures of Section~\ref{sec:experiments} are measured under those conditions.

\begin{figure}[pos=!ht]
\centering
\includegraphics[width=0.8\columnwidth]{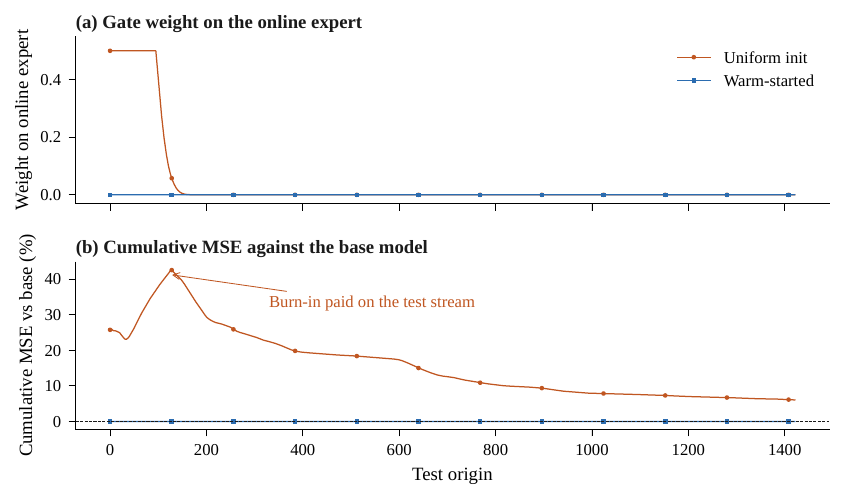}
\caption{Two-expert gate on Exchange with PatchTST, one run, from a uniform initialization and after the held-out warm start. (a) Weight on the online expert, which alone costs 65\% on this pair; no loss matures before origin 96. (b) Cumulative MSE relative to the base model; the uniform start ends 6.1\% above the base model and the warm start matches it from the first prediction.}
\label{fig:burnin}
\end{figure}

\paragraph{Warm start}
The regret bound is asymptotic, and on a short stream the descent from a uniform start is itself a cost. On Exchange with PatchTST, the gate holds half its weight on the online expert until the first losses mature at origin 96, although that expert alone costs 65\% on this pair (Table~\ref{tab:main}). It then reduces that weight to zero between origins 144 and 156 across the five runs. The stream still ends 5.8\% above the base model on the five-run mean, and 6.1\% above it on the run traced in Fig.~\ref{fig:burnin}. Our remedy replays the identical Hedge update, with the same $\eta$, normalization and maturation rule, over a dedicated warm slice of the held-out split. That slice holds $H + 200$ origins taken immediately before the early-stopping tail and is disjoint from both the correctors' fitting region and that tail. The final replayed weights initialize the test stream. Burn-in therefore happens on held-out data. On the same pair, the warm-started gate opens with $w \approx 0$ on that expert and matches the base model from the first prediction. The replay therefore acts as a data-dependent prior: where a corrector is harmful, the test stream opens with most of the weight already on the base model, the expert the downside is measured against. The same replay mechanism extends unchanged to any number of experts.

The layout constants are held fixed within each study, and none was changed after the load streams were opened. The benchmarks and the load streams share the warm slice of $H+200$ matured origins and the 10\% early-stopping tail. The tail width was chosen to keep a larger early-stopping sample, and panel B of Table~S13 varies it. The online corrector's update cadence is one optimizer step per 64 matured origins on the benchmarks and one per 8 on the load streams, whose matured history is two orders of magnitude shorter. Section~S7 perturbs these constants together with the trust-region radius $\delta$ of Section~\ref{sec:experts}, whose values Section~S1 lists. The gate's learning rate has its own sensitivity analysis (Section~\ref{sec:ablations}).

\subsection{Calibrated intervals via online conformal tracking}
\label{sec:conformal}

For uncertainty, we wrap the gated point forecast in an adaptive conformal layer, which reads the gate's output and feeds nothing back into it. One tracker per channel and horizon step maintains a radius $q$, and the interval at each origin is the point forecast plus and minus $q$. After every matured outcome the radius grows if the outcome fell outside the interval and shrinks if it fell inside. The update follows adaptive conformal inference \citep{gibbs2021}, but acts on the radius rather than on the miscoverage level, as in the quantile trackers of \citet{angelopoulos2023},
\begin{equation}
q \leftarrow q + \gamma\,(\mathrm{err} - \alpha), \qquad \mathrm{err} = \mathbb{1}\{\lvert y - \tilde{y} \rvert > q\}.
\label{eq:aci}
\end{equation}
The tracker is fed only by the maturation buffer, and its step size is $\gamma = 0.005$ throughout, never tuned per dataset. Recomputing it at $\gamma = 0.002$ and $\gamma = 0.01$ on the 14 trained-base-model pairs moves the mean Winkler score by at most 1.2\%, with the value used scoring best of the three, while mean 90\% coverage moves by about one point in either direction.

The guarantee inherited from \citet{gibbs2021} and \citet{angelopoulos2023} is that the realized coverage converges to $1 - \alpha$ over a long run, even under distribution shift. It says nothing about any finite stretch of the stream. The stronger finite-sample guarantee of conformal prediction requires exchangeable data, which a dependent series does not provide. The layer leaves the point forecast untouched, so the experiments can separate the contribution of the interval method from that of the correction. The adaptive tracker improves on static split calibration of the same forecast on every benchmark pair and in every load zone. Correcting the point forecast improves the intervals on most pairs (Section~\ref{sec:intervals}).

\section{Experiments}
\label{sec:experiments}

\subsection{Setup}
\label{sec:setup}

We evaluate on seven multivariate benchmarks: the four ETT series of \citet{zhou2021informer}, and the Weather, Electricity (ECL in the tables) and Exchange series from the collection assembled by \citet{wu2021autoformer}. These are the series on which the trained base models and the adaptation methods of Section~\ref{sec:baselines} were developed and report their results, so each frozen forecaster is used where it was designed to be used. The input length is 384 and the horizon 96. All splits are chronological, and Section~S1 gives their sizes. The eighth series of the same collection, weekly influenza-like illness (ILI), is too short to host the held-out layout and is treated in Section~\ref{sec:conditions}. The main results are at horizon 96, with horizons 192 and 336 in Section~S11. Because the warm slice needs $H + 200$ matured held-out origins, a dataset-model pair with a held-out split too short to host the slice is refused at that horizon (Section~\ref{sec:conditions}). We use four frozen base models. DLinear \citep{Zeng2023DLinear} and PatchTST \citep{Nie2023PatchTST} are trained for each dataset and run, then frozen. Chronos-Bolt Base (205M) \citep{chronos2024} and TimesFM 2.5 (200M) \citep{das2024timesfm} are zero-shot foundation models used without any fitting. Both checkpoints postdate the cited papers and have no papers of their own, so we cite the release documentation where a property is specific to the checkpoint \citep{chronosboltcard2024,timesfm25card2025}.

\paragraph{Runs and dispersion}
Every trained component is run five times, and every dispersion is the standard deviation over those runs, taken with denominator $n-1$ and printed after a $\pm$ sign. Worst-case statements at the pair or zone level refer to five-run means. The per-run cell is the stricter unit, and we report cell-level worst cases where a claim depends on them. A percentage change is computed from the run means of the two MSEs it compares, not as the mean of the five per-run changes. The two coincide whenever the base model's MSE is identical on every run, which is the case for the zero-shot and the statistical base models. They differ on DLinear and PatchTST, by up to 1.6 points on Exchange and by under 0.2 elsewhere. A dagger on an entry marks a change smaller in absolute value than the standard deviation of the five per-run changes, that is, a change within the run-to-run spread. Tables print that standard deviation beside the mean.

\paragraph{Experimental units}
A dataset-model pair, or pair, is one dataset with one base model, and a cell is one pair at one run. Each of the 28 pairs is run five times, which gives 140 cells. A cell's change is taken against the base model of the same run. The downside we report throughout is the largest such change over the pairs or cells a table names. It is measured on the runs, and is distinct from the regret bound (\ref{eq:regret}) of Section~\ref{sec:gate}.

\paragraph{Significance testing}
Where a test statistic is reported, it is the Diebold--Mariano test of equal predictive accuracy on the per-origin differences in squared error \citep{dm1995,harvey1997}. Comparisons of more than two methods use the rank test of Section~\ref{sec:ranks}. The Diebold--Mariano statistics are reported per run beside the effect sizes and are not adjusted for multiple comparisons; no claim rests on a single one of them. Those differences are autocorrelated, because the forecasts of consecutive origins share $H$ steps of outcome. We estimate their variance with a Newey--West estimator \citep{neweywest1987} at lag $H$, one lag beyond the order $H{-}1$ that \citet{harvey1997} derive for $H$-step-ahead errors. We standardize every series channel by channel, subtracting the training-split mean and dividing by the training-split standard deviation. Every error in the paper is reported on that standardized scale.

\paragraph{Methods compared}
Besides the frozen base model and the combination, we report three single correctors, each run alone with the gate removed. The \emph{static} corrector is the library's $E_1$, fitted once on the training split. The \emph{online} corrector is the library's $E_2$, fitted on the held-out fit region and then updated from matured outcomes during the stream. The \emph{held-out} corrector is not in the library. It has the static corrector's architecture, but it is fitted on the held-out fit region instead of the training split, and it never updates. The three differ on two axes: where the corrector is fitted, and whether it keeps learning during the stream (Table~S4).

\paragraph{Reproducibility}
Runs reproduce bit for bit within one computing environment; Section~S4 records the pinning and the environment scope.

\paragraph{Data-region separation}
No parameter or threshold used at test time is estimated on the test stream. Standardization constants come from the training split, the correctors' parameters from the training split and the held-out fit region, their stopping epoch from the early-stopping tail, and the gate's initial weights from the warm slice. Table~\ref{tab:data-usage} states which region decides what and which decisions each region is barred from. No region is used twice, and the test stream sets no parameter; the gate and the online corrector update their state there only from matured outcomes, by the rule fixed in advance. The layout constants of Section~\ref{sec:gate} and the gate's learning rate are held at one value across every dataset, base model, horizon and experiment, so no configuration is selected against a test result. The one exception is the two slowed learning rates of Section~\ref{sec:quasistatic}. They were chosen after test performance had been seen, and that section reports them as a diagnosis, not a configuration. The code and the result files behind every table are available as described under Data and code availability.

\begin{table}[pos=!ht]
\caption{Exact data-use specification: what each region decides, and what it never touches.}
\label{tab:data-usage}
\centering
\footnotesize
\begin{tabular}{lll}
\toprule
Region & Decides & Never touches  \\
\midrule
Training split & Base model and corrector parameters & Gate weights, evaluation  \\
Held-out fit region & Online corrector's initial fit & Gate weights, evaluation  \\
Held-out warm slice & Gate's initial weights, by replaying the update & Corrector parameters, evaluation  \\
Held-out early-stopping tail & Correctors' stopping epoch & Gate weights, evaluation  \\
Test stream & Gate and online-corrector state, from matured losses only & Any parameter or threshold  \\
\bottomrule
\end{tabular}
\end{table}

\subsection{Downside control on trained and foundation base models}
\label{sec:main-results}

\begin{figure}[pos=!ht]
\centering
\includegraphics[width=0.9\columnwidth]{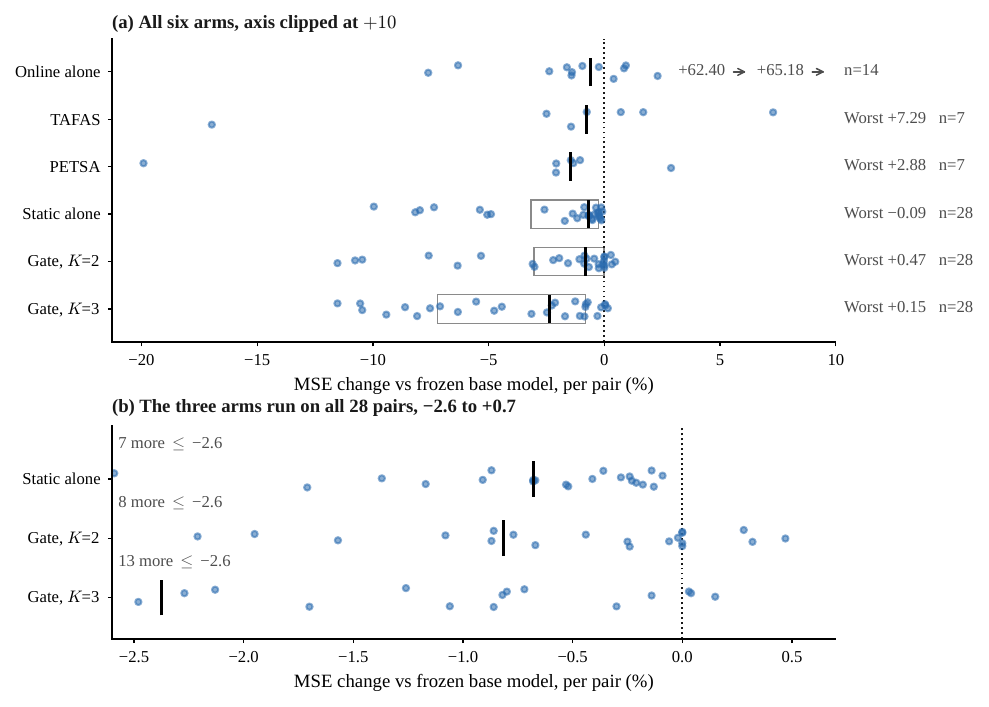}
\caption{MSE change against the frozen forecaster, one point per pair, five-run mean, median marked. Panel (a): each method on the pairs it was evaluated on, so $n$ differs by row. The static corrector and the two gates run on all 28 pairs, and a box behind their points marks the interquartile range across pairs. The online corrector alone runs only on the 14 pairs with a trained base model, and TAFAS and PETSA only on the seven DLinear pairs of Table~\ref{tab:fairness}, too few for a box. The axis is clipped at $+10$. Panel (b) enlarges the three 28-pair rows over $[-2.6, +0.7]$, and the count at a row's left edge gives its points below $-2.6$.}
\label{fig:cells}
\end{figure}

In Table~\ref{tab:main}, we report each method as a percentage change in MSE against its frozen base model. The static and online correctors are better in complementary regimes. The static corrector is the better of the two on Weather, Electricity and Exchange with both base models, where the error structure is visible in the training data and does not change. The online corrector is better on ETTm1 and ETTm2, where the structure shifts. Neither corrector dominates, and correction is not free. On Exchange with PatchTST, the online corrector costs 65.2\% and the held-out corrector 85.1\% against the base model, while the static corrector improves it by 7.4\%. The gate follows the better corrector in each situation and avoids the Exchange loss, without being told which one it is in. Fig.~\ref{fig:cells} shows the same pairs one by one, together with those of the other two base models. Of the 14 trained-base-model pairs, 13 improve. The fourteenth, ETTh2 with PatchTST, sits at $+0.03$\% against a run standard deviation of $0.09$, inside its run spread. In the table, 17 entries lie within their run spread and carry a dagger. The largest of them is the two-expert gate's $-0.86$\% on ETTh2 with DLinear.

\begin{table}[pos=!ht]
\caption{MSE change versus frozen base model in percent, computed from the run means of the two MSEs, all origins, printed as mean $\pm$ standard deviation of the five per-run changes with one more decimal than the mean. The held-out corrector has the static architecture and is not a library expert. $K{=}2$ is the base model with the online corrector; $K{=}3$ adds the static corrector. Negative is better, bold is the row's best, a dagger marks a change smaller in absolute value than the standard deviation of the five per-run changes.}
\label{tab:main}
\centering
\scriptsize\setlength{\tabcolsep}{3pt}
\begin{tabular}{llr@{\,$\pm$\,}lr@{\,$\pm$\,}lr@{\,$\pm$\,}lr@{\,$\pm$\,}lr@{\,$\pm$\,}l}
\toprule
Base model & Dataset & \multicolumn{2}{c}{Static} & \multicolumn{2}{c}{Online} & \multicolumn{2}{c}{Held-out} & \multicolumn{2}{c}{Gate, $K{=}2\nodg$} & \multicolumn{2}{c}{Gate, $K{=}3\nodg$}  \\
\midrule
DLinear & ETTh1 & $-0.53$ & $0.075$ & $-1.40$ & $1.250$ & $+0.20$ & $1.568^\dagger$ & $-1.95$ & $0.926$ & $\mathbf{-2.13}$ & $0.901\nodg$  \\
 & ETTh2 & $-0.09$ & $0.139^\dagger$ & $\mathbf{-1.62}$ & $0.889$ & $-0.76$ & $0.868^\dagger$ & $-0.86$ & $0.894^\dagger$ & $-0.72$ & $0.901^\dagger$  \\
 & ETTm1 & $-0.52$ & $0.049$ & $\mathbf{-1.42}$ & $0.873$ & $-0.16$ & $0.996^\dagger$ & $-0.25$ & $0.947^\dagger$ & $-0.30$ & $0.718^\dagger$  \\
 & ETTm2 & $-0.87$ & $0.146$ & $\mathbf{-7.61}$ & $3.265$ & $-7.46$ & $3.216$ & $-7.59$ & $3.274$ & $-7.53$ & $3.347\nodg$  \\
 & Weather & $\mathbf{-4.90}$ & $1.220$ & $+0.40$ & $1.029^\dagger$ & $+5.20$ & $1.470$ & $-0.24$ & $0.550^\dagger$ & $-4.76$ & $1.216\nodg$  \\
 & ECL & $\mathbf{-2.59}$ & $0.936$ & $-0.95$ & $0.598$ & $-0.64$ & $0.520$ & $-1.08$ & $0.595$ & $-2.48$ & $0.931\nodg$  \\
 & Exchange & $\mathbf{-9.96}$ & $5.645$ & $+62.40$ & $12.751$ & $+83.74$ & $13.834$ & $+0.00$ & $0.0001^\dagger$ & $-9.42$ & $5.723\nodg$  \\
 & \textit{Worst} & \multicolumn{2}{r}{$-0.09\nodg$} & \multicolumn{2}{r}{$+62.40\nodg$} & \multicolumn{2}{r}{$+83.74\nodg$} & \multicolumn{2}{r}{$+0.00\nodg$} & \multicolumn{2}{r}{$\mathbf{-0.30}\nodg$}  \\
\midrule
PatchTST & ETTh1 & $-0.13$ & $0.033$ & $+2.30$ & $0.238$ & $+4.71$ & $0.718$ & $-0.02$ & $0.100^\dagger$ & $\mathbf{-0.14}$ & $0.047\nodg$  \\
 & ETTh2 & $\mathbf{-0.18}$ & $0.094$ & $+0.93$ & $0.759$ & $+1.21$ & $0.713$ & $+0.28$ & $0.085$ & $+0.03$ & $0.085^\dagger$  \\
 & ETTm1 & $-0.24$ & $0.143$ & $\mathbf{-2.38}$ & $1.340$ & $-1.41$ & $1.403$ & $-2.21$ & $1.604$ & $-2.27$ & $1.473\nodg$  \\
 & ETTm2 & $-0.41$ & $0.101$ & $-6.32$ & $3.289$ & $-5.82$ & $3.426$ & $\mathbf{-6.34}$ & $3.286$ & $-6.33$ & $3.284\nodg$  \\
 & Weather & $\mathbf{-1.37}$ & $0.665$ & $+0.85$ & $0.292$ & $+4.00$ & $0.661$ & $-0.06$ & $0.120^\dagger$ & $-1.26$ & $0.709\nodg$  \\
 & ECL & $\mathbf{-1.17}$ & $0.172$ & $-0.24$ & $0.258^\dagger$ & $+0.17$ & $0.274^\dagger$ & $-0.44$ & $0.265$ & $-1.06$ & $0.153\nodg$  \\
 & Exchange & $\mathbf{-7.36}$ & $1.391$ & $+65.18$ & $8.535$ & $+85.07$ & $9.642$ & $+0.00$ & $0.0001^\dagger$ & $-7.10$ & $1.445\nodg$  \\
 & \textit{Worst} & \multicolumn{2}{r}{$\mathbf{-0.13}\nodg$} & \multicolumn{2}{r}{$+65.18\nodg$} & \multicolumn{2}{r}{$+85.07\nodg$} & \multicolumn{2}{r}{$+0.28\nodg$} & \multicolumn{2}{r}{$+0.03\nodg$}  \\
\bottomrule
\end{tabular}
\end{table}

\paragraph{The static corrector alone}
By the worst-case criterion, the static corrector is the safest single method on the benchmarks. It deteriorates on no pair. Its worst change over the 28 pairs is $-0.09$\%, against $+0.15$ for the gate, and the rank test does not separate the two (Table~\ref{tab:mcb}). Its mean change is $-2.24$\% against $-3.99$ for the gate, and the same order holds cell by cell over the 140 pair-run cells. The static corrector cannot follow error structure that moves. On ETTm2, it gains under one percent on both base models, while the gate gains between six and eight percent.

On the frozen statistical base models of Section~\ref{sec:ranks}, the static corrector run alone improves the eight rows by at most 13.4\%, while the combination improves them by 21.4 to 91.3\% (Table~S12). The gate gets there by placing almost all of its weight on the online expert, which no static corrector can do. The static corrector improves every benchmark pair and every statistical-base-model row. It stops short where the base model is weak enough for correction to become replacement: on the statistical base models, on ETTm1 and ETTm2 with the foundation models (Table~\ref{tab:fm}) and on Hungary in Section~\ref{sec:opsd}. The gate also improves the datasets whose error structure shifts. Its worst pair sits 0.15\% above the base model, while the static corrector's worst pair sits 0.09\% below it. Over the 140 pair-run cells, the stricter unit, the largest deterioration is $+0.46$\% for the gate and $+0.08$\% for the static corrector (panel A of Table~S7).

Which method a practitioner should prefer therefore depends on whether the error structure is stable. The gate learns this from matured losses, so the practitioner need not know it in advance. Section~\ref{sec:opsd} returns to the question with a forecaster whose error structure differs by zone, and there the static corrector deteriorates for the first time.

Table~S10 shows that the gate does not simply average. The expert receiving the largest mean weight matches the regime on six of the seven datasets. The online corrector leads on ETTm2, and the static corrector leads on Weather, Electricity and Exchange, in each case with weights up to 0.998. Mixtures appear on ETTh1, ETTh2 and ETTm1, where neither expert is clearly better; there the static corrector holds the larger share.

\paragraph{Equal, fixed and adaptive weights}
Panel A of Table~S7 adds two reference points beside the static corrector and the gate, on the same basis as Table~\ref{tab:main}. The equal-weight average of the three library experts is the benchmark the combination literature asks any weighting scheme to clear \citep{clemen1989,wang2023combinations}. Here it improves 24 of 28 pairs at a mean of $-1.54$\%, but it costs $+9.06$\% on its worst pair, Exchange with PatchTST, because it gives a harmful online corrector a third of the weight. Two corrections of the intercept type read the same matured error sequence as the online corrector and fit nothing. The intercept correction adds the mean of the most recent 200 matured errors at the same lead \citep{clements1996intercept,castle2015robust}. The exponentially weighted correction instead adds an exponentially weighted mean of those errors with a half-life of 100 origins. Neither improves a single benchmark pair, and both cost more than 13\% on average and up to $+33.1$\%. The benchmark errors therefore carry no bias that a running mean can remove, and the gate's gain on these series is not bias removal. Section~S5 gives every pair. Section~\ref{sec:opsd} returns to both correctors on the load data, where the intercept correction is strong and the ordering reverses.

The error second moments stored for every combination cell give the loss of any fixed weights exactly, including the equal average. For a fixed weight vector $w$, the loss is
\begin{equation}
L(w) = w^\top M w, \qquad M_{jk} = \frac{1}{|\mathcal{T}|} \sum_{o \in \mathcal{T}} \langle e^{(j)}_o, e^{(k)}_o \rangle / (HC).
\label{eq:fixedw}
\end{equation}
Here $M$ is the matrix of raw error second moments over the matured test origins $\mathcal{T}$, and $e^{(k)}_o = y_o - y^{(k)}_o$. Panel B of Table~S7 evaluates three fixed rules and one error-driven rule under (\ref{eq:fixedw}) against the gate on the same cells. Equal weighting is good on average and has no bound on its worst case. It lowers the mean error on most pairs. It costs 9.7\% on the pair where one expert is bad, and 17.1\% when the base model is dropped from the average. On the load data, it improves four of the seven zones against the gate's seven, because an expert that costs over 100\% on a zone enters the average at full weight.

The rule most often used in the combination literature sets each weight in proportion to the inverse of recent error \citep{bates1969}. We recompute it on the matured stream over a trailing window. It improves on the equal average, lowering the mean error and cutting the worst pair from 9.7 to 2.8\%. It stays well behind the gate on both axes, on the matured-origin basis of panel B of Table~S7: it gives up more than a point and a half of mean improvement and more than two points on the worst pair. The ordering is the same on the load data. The weights it produces are near-uniform on almost every cell. On these streams, a rule based on the size of recent errors cannot separate experts whose errors are similar in size but differently distributed. It therefore collapses toward the average it is meant to improve on. It moves away from equal weights only where one expert is clearly better, as on Hungary and Germany. Shortening the window adds estimation noise without improving that separation.

\begin{figure}[pos=!ht]
\centering
\includegraphics[width=0.95\columnwidth]{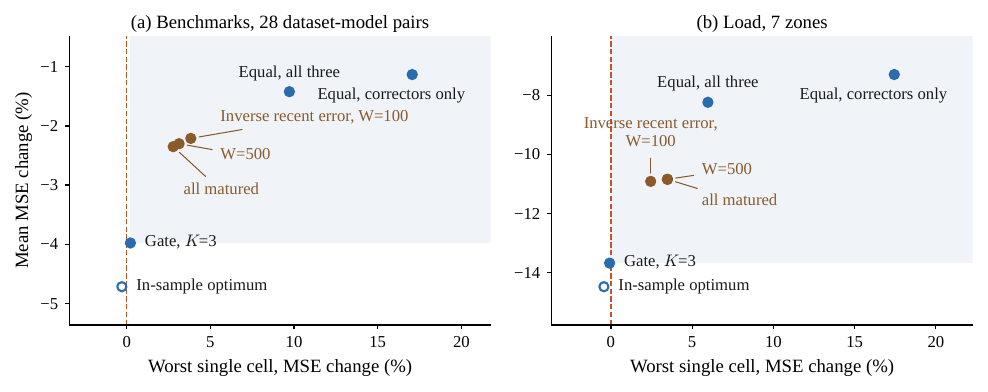}
\caption{Mean improvement against worst cell for the rules of panel B of Table~S7, on its matured-origin basis. Shading marks the region the three-expert gate dominates; only the in-sample optimum, an oracle, lies outside that region.}
\label{fig:pareto}
\end{figure}

The in-sample optimum is fitted on the same stream it is scored on, so it is an oracle, not a method. It improves on the gate by a median of 0.8 percentage points, which bounds what any better-chosen fixed weight could have recovered. Fig.~\ref{fig:pareto} places the methods on the two axes that the design objective names, mean improvement and worst cell.

\paragraph{Error correlation and the cost of the realized weight}
On 415 combination runs from every stage of the study, we compare the in-sample optimal weight between the two correcting experts with the weight the gate realizes. The combination variance is evaluated at both (Section~S5). Where the experts are most alike, the optimal weight is both hardest to estimate and least consequential. This is the regime in which trimming to the simplex is predicted to do best \citep{radchenko2023similar}, and in which the weight-estimation variance identified by \citet{claeskens2016puzzle} is largest. The gate never estimates that weight, and gives up a fraction of a point by not knowing it.

\begin{table}[pos=!ht]
\caption{The static corrector alone and the combination on frozen zero-shot foundation models. MSE change versus base model in percent, mean $\pm$ standard deviation over five runs, all origins; the standard deviation carries one more decimal than the mean so that no entry rounds to zero. A dagger marks a change smaller in absolute value than that standard deviation.}
\label{tab:fm}
\centering
\footnotesize
\begin{tabular}{lr@{\,$\pm$\,}lr@{\,$\pm$\,}lr@{\,$\pm$\,}lr@{\,$\pm$\,}l}
\toprule
 & \multicolumn{4}{c}{Chronos-Bolt} & \multicolumn{4}{c}{TimesFM}  \\
\cmidrule(lr){2-5}\cmidrule(lr){6-9}
Dataset & \multicolumn{2}{c}{Static} & \multicolumn{2}{c}{Gate, $K{=}3$} & \multicolumn{2}{c}{Static} & \multicolumn{2}{c}{Gate, $K{=}3$}  \\
\midrule
ETTh1 & $-0.23$ & $0.001$ & $-0.80$ & $0.265$ & $-0.28$ & $0.003$ & $+0.04$ & $0.062^\dagger$  \\
ETTh2 & $-0.21$ & $0.049$ & $+0.15$ & $0.081$ & $-0.14$ & $0.004$ & $-0.86$ & $0.628\nodg$  \\
ETTm1 & $-0.68$ & $0.001$ & $-10.55$ & $0.288$ & $-0.36$ & $0.007$ & $-3.15$ & $0.546\nodg$  \\
ETTm2 & $-0.68$ & $0.009$ & $-11.53$ & $0.217$ & $-0.67$ & $0.025$ & $-10.46$ & $0.217\nodg$  \\
Weather & $-7.97$ & $0.104$ & $-8.61$ & $0.464$ & $-5.38$ & $0.057$ & $-5.54$ & $0.054\nodg$  \\
ECL & $-0.91$ & $0.076$ & $-0.82$ & $0.068$ & $-1.71$ & $0.182$ & $-1.70$ & $0.079\nodg$  \\
Exchange & $-8.17$ & $0.212$ & $-8.09$ & $0.213$ & $-5.06$ & $0.289$ & $-4.43$ & $0.301\nodg$  \\
\midrule
\textit{Worst} & \multicolumn{2}{r}{$-0.21$\hphantom{\,$\pm$\,0.001}} & \multicolumn{2}{r}{$+0.15$\hphantom{\,$\pm$\,0.265}} & \multicolumn{2}{r}{$-0.14$\hphantom{\,$\pm$\,0.003}} & \multicolumn{2}{r}{$+0.04$\hphantom{\,$\pm$\,0.062}}  \\
\bottomrule
\end{tabular}
\end{table}

\paragraph{Transfer to frozen foundation models}
\label{sec:fm}

We fit neither Chronos-Bolt nor TimesFM. The correctors' architecture and hyperparameters, the gate's learning rate and its normalization, and the warm-start protocol were all fixed on the trained base models. Each foundation model enters through the same interface as any other frozen forecast, with nothing adjusted for it. Table~\ref{tab:fm} reports the combination against both. The static corrector alone improves all 14 pairs and leads the combination on six of them, by at most 0.63 points. On ETTm1 and ETTm2, it stays under one percent while the combination reaches between three and twelve percent, the same division of labor as on the trained base models. Gains reach 11.5\% on Chronos-Bolt and 10.5\% on TimesFM. The worst deteriorations are 0.15 and 0.04\%, and the first is the worst over all 28 pairs in the study. The two deteriorations differ. The 0.15 on ETTh2 is small but exceeds its run standard deviation of 0.08, while the 0.04 on ETTh1 lies inside its own spread.

The two models are pretrained independently, on different corpora and with different architectures, so the downside control observed on both is not specific to one foundation model. Whether either pretraining corpus overlaps the public benchmarks does not affect the operation of the layer, which reads the frozen forecast and the matured outcome and nothing else, though it would bear on how the zero-shot accuracy itself is read. The load study of Section~\ref{sec:opsd} scores it on outcomes settled months after the forecast was issued, which no pretraining corpus contains.

\subsection{Comparison with test-time adaptation methods}
\label{sec:baselines}

In Table~\ref{tab:fairness}, we compare against TAFAS \citep{tafas2025} and PETSA \citep{petsa2025} on DLinear. Both are re-run under the standard split with our frozen base models, so these numbers differ from the published ones. Their PatchTST implementations are built into their adaptation code, so using our frozen weights would require editing their methods. We therefore record those pairs as incompatible.

\begin{figure}[pos=!ht]
\centering
\includegraphics[width=0.9\columnwidth]{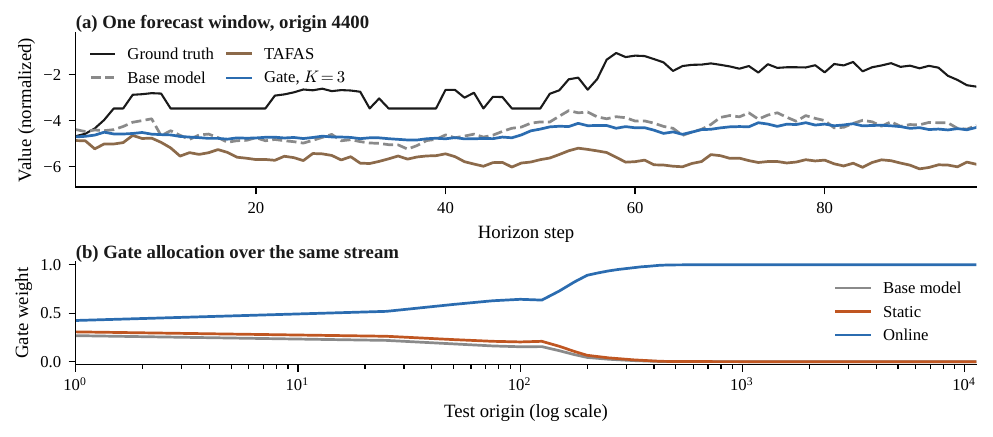}
\caption{Adaptation that recalibrates a forecaster's input and output can move a forecast away from the truth. One window and one stream of ETTm2 with DLinear, the dataset on which TAFAS deteriorates most in Table~\ref{tab:fairness}.}
\label{fig:hurts}
\end{figure}

By MSE alone, the table is mixed: both baselines have lower MSE than the combination on three of the seven pairs. The accounting columns show why the comparison is not purely one of accuracy. All three methods leave the base model frozen and update add-on parameters, but at very different scales. TAFAS calibrates the input and the output of the forecaster with a temporal map per channel, which reaches 50.4 million parameters on Electricity. It deteriorates the base model on three of seven datasets, by as much as 7.3\%. Fig.~\ref{fig:hurts} shows what that deterioration looks like inside a single window. PETSA updates add-on parameters at a scale comparable to ours but still deteriorates Electricity by 2.9\%. The combination deteriorates no dataset and updates 37{,}252 add-on parameters, three orders of magnitude fewer than TAFAS. It is also the only one of the three that supplies intervals.

Panel (a) of Fig.~\ref{fig:hurts} follows one window of one ETTm2 channel, the one named MULL in the source file. The ground truth rises while TAFAS drifts downward and ends 2.8 times worse than the frozen base model it started from. The gated forecast stays close to that base model. Panel (b) follows the gate on the same stream. The gate concentrates on the online expert within a few hundred origins and stays there. Both panels are drawn from a single run, so their values differ from the five-run means of Table~\ref{tab:fairness}.

\begin{table}[pos=!ht]
\caption{Accounted comparison on DLinear. MSE change versus base model in percent, computed from the run means of the two MSEs, printed as mean $\pm$ standard deviation of the five per-run changes with one more decimal than the mean; update surface and outcome assumption are part of the comparison.}
\label{tab:fairness}
\centering
\footnotesize
\begin{tabular}{lr@{\,$\pm$\,}lr@{\,$\pm$\,}lr@{\,$\pm$\,}l}
\toprule
Dataset & \multicolumn{2}{c}{TAFAS} & \multicolumn{2}{c}{PETSA} & \multicolumn{2}{c}{Gate, $K{=}3$}  \\
\midrule
ETTh1 & $-1.44$ & $1.488$ & $-1.34$ & $1.350$ & $\mathbf{-2.13}$ & $0.901\nodg$  \\
ETTh2 & $\mathbf{-2.50}$ & $2.125$ & $-2.08$ & $2.077$ & $-0.72$ & $0.901\nodg$  \\
ETTm1 & $-0.76$ & $0.703$ & $\mathbf{-1.05}$ & $0.325$ & $-0.30$ & $0.718\nodg$  \\
ETTm2 & $+7.29$ & $4.171$ & $-2.09$ & $3.383$ & $\mathbf{-7.53}$ & $3.347\nodg$  \\
Weather & $+0.71$ & $1.919$ & $-1.45$ & $1.458$ & $\mathbf{-4.76}$ & $1.216\nodg$  \\
ECL & $+1.68$ & $0.504$ & $+2.88$ & $0.530$ & $\mathbf{-2.48}$ & $0.931\nodg$  \\
Exchange & $-16.96$ & $12.262$ & $\mathbf{-19.91}$ & $11.915$ & $-9.42$ & $5.723\nodg$  \\
\midrule
\textit{Worst} & \multicolumn{2}{r}{$+7.29$} & \multicolumn{2}{r}{$+2.88$} & \multicolumn{2}{r}{$\mathbf{-0.30}$}  \\
Update surface & \multicolumn{2}{r}{Add-on} & \multicolumn{2}{r}{Add-on} & \multicolumn{2}{r}{Add-on}  \\
Params updated & \multicolumn{2}{r}{1.1M--50.4M} & \multicolumn{2}{r}{64.8K--2.6M} & \multicolumn{2}{r}{37.3K}  \\
Outcome protocol & \multicolumn{2}{r}{Partial} & \multicolumn{2}{r}{Partial + matured} & \multicolumn{2}{r}{Matured}  \\
Intervals & \multicolumn{2}{r}{No} & \multicolumn{2}{r}{No} & \multicolumn{2}{r}{Yes}  \\
\bottomrule
\end{tabular}
\end{table}

Part of the remaining accuracy gap comes from the outcome protocol, not from architecture. We ran PETSA's calibration module under our matured-outcome rule, with its code unmodified and only its partial-outcome update path left uncalled. Its Exchange gain then falls from 19.9 to 3.3\%. Partial outcomes are a legitimate design choice where a horizon can be scored before it completes, and this measurement is not a criticism of it. \citet{deltaadapter2026} re-run several adaptation methods under a leakage-free protocol and report the same direction of effect. A comparison that does not state the outcome assumption mixes the corrector's gains with the protocol's.

\subsection{Ranks, scale and statistical base models}
\label{sec:ranks}

Comparing more than two methods over many datasets calls for a rank test, not a sequence of pairwise ones. Table~\ref{tab:mcb} reports Friedman mean ranks with the Nemenyi critical distance \citep{demsar2006} on three complete block sets. We report each set separately because not every method runs on the same pairs. The 28 blocks are not independent series: four base models share each of the seven datasets, so the test reads on this set of configurations. Rank tests of this kind entered forecast evaluation with \citet{koning2005m3}, who compared each method in the M3 competition against the best and against the mean. The Nemenyi distance used here compares every pair instead. On the 28 pairs, the test is significant at the 5\% level. It separates the frozen base model from every other method, and no other two methods from each other. The combination ranks first at 1.786 and the static corrector follows at 1.893, a lead of about a tenth of a rank against a critical distance of 0.886. On the 14 trained-base-model pairs, the lead narrows further against a critical distance of 1.630. Because the 28 blocks share seven datasets, the same test with the dataset as the block, seven blocks in all, is the stricter reading (Table~S23): it still rejects at the 5\% level on both sets ($p = 0.008$ and $p = 0.038$), and the one separation that survives the wider critical distance is the three-expert gate against the frozen base model.

Mean change and the count of improved pairs order the combination and the static corrector in opposite ways. The static corrector improves every one of the 28 pairs. The combination deteriorates on three of them but gains more on average. Rank, mean, count and worst case answer different questions. On the 14 trained-base-model pairs, the online corrector is also reported on its own. It ranks between the base model and the other correctors, although its worst pair is $+65.18$\%. The two-expert gate's worst pair, $+0.47$\%, is ETTh2 with Chronos-Bolt, where the three-expert gate sits at $+0.15$. Table~S9 gives the two-expert gate on all 14 foundation-model pairs.

Seven datasets cannot resolve the comparison with the two adaptation methods at all. The test does not reject at the 5\% level. The critical distance of 3.405 also exceeds the whole range of mean ranks, so no two methods in that block set are separated.

\begin{table}[pos=!ht]
\caption{Friedman mean ranks with the Nemenyi critical distance (CD), mean MSE change in percent and pairs improved, all origins. Worst is the largest deterioration over the block set, in percent.}
\label{tab:mcb}
\centering
\footnotesize
\begin{tabular}{llrrrr}
\toprule
Block set & Method & Mean rank & Mean & Improved & Worst  \\
\midrule
28 pairs & Frozen base model & 3.661 & $0.00$ & n/a & $+0.00$  \\
$\mathrm{CD}=0.886$ & Static & 1.893 & $-2.24$ & 28/28 & $-0.09$  \\
$p<0.001$ & Gate, $K{=}2$ & 2.661 & $-2.43$ & 21/28 & $+0.47$  \\
 & Gate, $K{=}3$ & \textbf{1.786} & $\mathbf{-3.99}$ & 25/28 & $+0.15$  \\
\midrule
14 trained & Frozen base model & 4.286 & $0.00$ & n/a & $+0.00$  \\
$\mathrm{CD}=1.630$ & Static & 2.214 & $-2.17$ & 14/14 & $-0.09$  \\
$p=0.001$ & Online & 3.429 & $+7.86$ & 8/14 & $+65.18$  \\
 & Gate, $K{=}2$ & 2.929 & $-1.48$ & 11/14 & $+0.28$  \\
 & Gate, $K{=}3$ & \textbf{2.143} & $\mathbf{-3.25}$ & 13/14 & $+0.03$  \\
\midrule
7 DLinear & Frozen base model & 6.000 & $0.00$ & n/a & $+0.00$  \\
$\mathrm{CD}=3.405$ & Static & 3.714 & $-2.78$ & 7/7 & $-0.09$  \\
$p=0.25$ & Online & 3.714 & $+7.11$ & 5/7 & $+62.40$  \\
 & Gate, $K{=}2$ & 3.857 & $-1.71$ & 6/7 & $+0.00$  \\
 & Gate, $K{=}3$ & 3.143 & $\mathbf{-3.91}$ & 7/7 & $-0.30$  \\
 & TAFAS & 4.143 & $-1.71$ & 4/7 & $+7.29$  \\
 & PETSA & 3.429 & $-3.58$ & 6/7 & $+2.88$  \\
\bottomrule
\end{tabular}
\end{table}

\paragraph{Scale-free measures}
A percentage change against a frozen forecaster says nothing about how hard the series is. Table~S11 reports the mean absolute scaled error (MASE) and the root mean squared scaled error (RMSSE) in their seasonal form. These are the scaled errors of \citet{hyndman2006mase} and \citet{makridakis2022m5}, whose denominator is the error of a naive forecaster on the training data. We replace the one-step naive forecaster with a seasonal one at each dataset's calendar period $m$,
\begin{equation}
\mathrm{MASE} = \frac{\mathrm{mean}\,\lvert y - \tilde{y} \rvert}{\mathrm{mean}_{\,t}\,\lvert y_t - y_{t-m} \rvert}, \qquad \mathrm{RMSSE} = \sqrt{\frac{\mathrm{mean}\,(y - \tilde{y})^2}{\mathrm{mean}_{\,t}\,(y_t - y_{t-m})^2}},
\label{eq:scalefree}
\end{equation}
with the denominators taken over the training split.

The combination lowers MASE on 13 of the 14 pairs. Weather with PatchTST is the exception. There the squared error falls 1.26\% while the absolute error rises 1.81\%, so a measure built on absolute error rises with it. The static corrector moves the same way on that pair, which places the divergence in the data rather than in the gate. RMSSE follows squared error. On ETTh2 with PatchTST, the one pair of the 14 whose squared error rises, the RMSSE rise is by construction about half of the MSE change. It is smaller than the printed precision, and Table~\ref{tab:main} marks the MSE change as inside its run spread.

On ETTh2, every method, including the base model, has MASE above one. The forecasters there are worse than a seasonal rule, so the combination is improving a forecast that a practitioner would not deploy. On Exchange, the period is one, so the reference is a random walk. Every MASE there is above seven except the static corrector's 6.99 with DLinear. The gains that Table~\ref{tab:main} reports there correct a forecast that is far from competitive; they do not improve a good one.

\paragraph{Statistical base models}
We also apply the same layer to frozen statistical base models, which shows what the size of the gain depends on. Table~S12 reports a seasonal naive forecaster and exponential smoothing on the four datasets whose calendar-day period is unambiguous at their sampling rate. Section~S6 gives the construction and the period convention. The improvement ranges from 21.4 to 91.3\%, an order of magnitude above what the trained base models allow, and none of the 120 corrected cells (eight rows, three corrected methods, five runs) deteriorates. The static corrector alone improves every row but stops 21 to 81 points short of the combination. On these base models, the gate selects rather than combines. Its mean weight on the online expert is 0.989 or above on every row, with essentially nothing on the frozen forecaster. The simplex permits this behavior, and the trained-base-model pairs never call for it.

\subsection{Intervals and ablations}
\label{sec:intervals}
\label{sec:ablations}

On the 14 trained-base-model pairs, we separate the contribution of the interval layer from that of the correction with three arms: the adaptive tracker on the gated forecast, and split calibration and the same tracker on the uncorrected forecast. The load study is treated in Section~\ref{sec:opsd-intervals}. All three are scored on all origins at the 90\% level by the Winkler score \citep{winkler1972}, the interval width plus a penalty for each observation outside it. Scores are five-run means per pair. To isolate the interval method, we compare the two calibrations on the same uncorrected point forecasts. The adaptive tracker scores better than split calibration on all 14 pairs. To isolate the correction, we keep the adaptive tracker fixed. The gated forecast is then no worse on 10 of 14 pairs, with the two Exchange pairs tied at the printed precision. It is worse on ETTh1 with PatchTST, on ETTh2 with both base models and on ETTm1 with DLinear. The full layer scores better than the split-calibrated frozen forecast on 13 of 14 pairs. Both contribute, and the adaptive layer does so more consistently than the correction.

Table~S14 compares the tracker with the native quantile head of Chronos-Bolt at a matched nominal level. The tracker is applied to the uncorrected Chronos-Bolt forecast, so only the interval method differs. Chronos-Bolt is trained on quantile levels 0.1 to 0.9 \citep{chronosboltcard2024}, so its widest native interval is 80\% and both rows use $\alpha = 0.2$. The native head is narrower and under-covers on every dataset. Its mean absolute coverage error is 0.056, against 0.020 for the tracker, which is better calibrated on six of seven datasets. The ordering holds at every nominal level (Fig.~S1). The better calibration comes with wider intervals.

The mean pinball loss over the nine deciles averages the quantile loss over quantile levels. Energy forecasting competitions use it to approximate the continuous ranked probability score \citep{hong2016gefcom}. By that measure the native head is better on six of seven datasets, 0.112 against 0.116. The tracker is better only on Exchange, where the native head under-covers most. Unconditional and conditional coverage tests \citep{kupiec1995,christoffersen1998} are reported in the result files for every cell. Pooled over origins, steps and channels, these tests use more than a million points per cell. They reject for both methods at every cell, so they do not separate the two. The adaptive layer is the better calibrated of the two in these experiments, while the native head has the lower mean pinball loss on six of the seven datasets.

\paragraph{Warm start}
Without the warm start, the gate opens at uniform weights and pays its burn-in on the test stream. On Exchange with PatchTST, the two-expert gate then costs 5.78\% against the base model on the five-run mean, the burn-in shown in Fig.~\ref{fig:burnin}. With the warm start, the gate matches the base model exactly. That unwarmed descent is the largest deterioration against a base model observed at horizon 96 with the reported library. Two of the three failures behind the applicability conditions of Section~\ref{sec:conditions} arise outside that scope: one from adding a fast fourth expert, and one at horizon 192. The third comes from scoring on a different outcome version, not from a deterioration against the outcome the gate learned from. The warm-started gate is conservative where correction does not help. On Weather with DLinear, the online corrector alone is harmful and the two-expert gate holds the pair near the base model. Adding the static expert then takes the combination to $-4.76$\%.

\paragraph{Learning rate of the gate}
Over a sixfold range of $\eta$, the largest MSE spread on any cell is 1.14\%. A larger $\eta$ pushes the allocation toward putting all the weight on one expert, without changing which expert that is. The warm start already lands close to such an allocation, which absorbs most of the difference. We fix $\eta = 0.1$ throughout, and panel C of Table~S13 gives the sweep.

\paragraph{Layout constants} The three layout constants of Section~\ref{sec:gate} and the trust-region radius are held fixed throughout. A dependence on them would show on the worst pair, ETTh2 with Chronos-Bolt at $+0.15$\%. Panel A of Table~S13 halves and doubles each one in turn on that pair, at five runs from the cached base-model forecasts. Two of the twelve points leave the pair worse than the reported configuration: a halved trust-region radius and a doubled update interval. Both differ from it by less than a twentieth of a point. Three points turn the deterioration into a gain: the halved warm slice and both alternative tail values. The reported setting is not the best point on its own grid. The early-stopping tail sets how much held-out data the correctors' stopping epoch is chosen on, so panel B of Table~S13 runs its three values on all 14 foundation-model pairs. A 5\% tail improves every one of them and turns the worst case into a gain of $0.17$\%, which would make the worst case reported here smaller. A 20\% tail costs Exchange with TimesFM its $4.4$\% gain and leaves it at $+0.56$\%, the largest deterioration anywhere in the sweep.

\section{Load forecasts with revised outcomes}
\label{sec:opsd}

The benchmarks above use base models that we trained and could in principle retrain. The same layer applies to a forecaster that cannot be retrained by anyone outside the organization that issues it. That forecaster is a day-ahead load forecast published by a transmission system operator (TSO), the company that runs a country's high-voltage grid and balances supply and demand on it. Each TSO forecasts the load of its bidding zone, the area within which electricity trades at a single price, for the following day. It publishes that forecast, together with the load actually measured, through the European Network of Transmission System Operators for Electricity (ENTSO-E). We call it the TSO forecast throughout. Its parameters are not published, so no user outside the operator can retrain it. Non-retrainability here is a property of the issuing organization, not of an interface.

The outcomes arrive in two versions at two speeds, as macroeconomic aggregates do in real-time data sets \citep{croushore2001realtime}. The Open Power System Data time series package of 2019-06-05 \citep{opsd2019} carries three series for each of 36 bidding zones: the TSO forecast and the actual load in two outcome versions. The provisional outcome is the load reported to the ENTSO-E Transparency Platform within one hour of the operating period, as archived in the data package we use. The settled outcome is the load as published in ENTSO-E Power Statistics up to three months later, after re-metering. The two differ by a zone-specific amount, from 0.6\% mean absolute difference in Hungary to 9.4\% in Italy (Table~S18). \citet{hirth2018entsoe} document the same split for 2015 and 2016: Transparency Platform load, delivered one hour after real time, deviates significantly and persistently from Power Statistics, which undergo revisions, and in most countries it is the smaller of the two. The difference reflects measurement as well as timing, because the Transparency column in the package holds the value that stood on the platform when the package was built. Either way, an operator who wants to learn from realized load while it is fresh learns from the provisional outcome and is eventually judged against the settled one.

The quality of these forecasts has been measured directly. \citet{kazmi2022tso} analyse five years of TSO load, wind and solar forecasts for 16 European countries from the Transparency Platform. Most of the forecasts beat a daily naive baseline, but in every country their errors remain strongly autocorrelated, and the authors conclude that the structure left in the residuals can be used to improve them. Published TSO forecasts have also been improved from their own history. \citet{mobius2025load} correct the ENTSO-E day-ahead load forecast in real time from the history of its error alone, and use the corrected series as an input to a price model. \citet{girolimetto2025terna} combine Terna's zonal forecasts with a daily naive forecast by stacked regression. Both show that the TSO forecast carries recoverable error and that its own history suffices to recover part of it. We ask two further questions: whether the recovery can avoid deteriorating any single zone, and whether it survives the revision of the outcome it was learned from.

\subsection{Data, zones and protocols}
\label{sec:opsd-data}

We fixed two zone screens before inspecting any zone outside a five-zone pilot set. We added a third screen after a forecast defect surfaced in one zone, and Section~S9 records the order. A zone is admitted if it has under 1\% missing values in every column, no gap longer than three hours in either outcome series, and no hour whose TSO forecast exceeds three times the load measured at the same time. None of the three screens looks at how any method performs: a zone enters on the completeness and plausibility of its published series alone. Seven zones pass: Germany (DE), Hungary (HU), Portugal (PT), Croatia (HR), Denmark (DK), Italy (IT) and Belgium (BE). The tables use these codes. The first five are the pilot zones, and Italy and Belgium entered when the selection was widened to all 36. Of the 29 rejections, 19 fall to the missing-fraction screen, eight to the gap screen and two to the forecast defect.

The evaluation window ends on 2019-01-31 rather than at the package end of 2019-04-30, because the settled outcome does not reach the later date in any zone. The publication delay thus appears as a coverage boundary. The forecast screen rejects the Netherlands, where the TSO forecast exceeds 500 GW in 96 hours against a load of 13.5 GW. \citet{kazmi2022tso} also flag the Dutch day-ahead load forecast on the platform as a likely data-quality problem rather than a forecasting one. For that one zone, the screen is post-hoc. Section~S9 lists every zone with its reason and the order in which the screens were written.

\label{sec:opsd-protocols}
The evaluation issues one origin per day at 00:00 UTC with a horizon of 24 hours. The TSO forecast for that day is the frozen base model $E_0$, and the correctors get a look-back of 168 hours. Each zone is corrected and gated independently, with its own correctors and its own weights. No cross-zone information enters any component at any point. The correctors read only the load history and the TSO forecast, with no weather or calendar covariate. The forecast being corrected already uses whatever covariates its issuer chose. The layer needs none of them and adds nothing the operator does not already have. We report the Netherlands, which the quality screen of Section~S9 excludes, as a sensitivity below. The protocol uses the three-way disjoint held-out layout of Section~\ref{sec:gate} and the maturation rule of Section~\ref{sec:protocol}, with five runs.

We standardize each zone with one mean and one standard deviation, fitted on the training hours of the provisional outcome. The forecast and both outcome versions share this scaling, so all three stay comparable.

We then cross learning and scoring with the two outcome versions, in three protocols (L1, L2 and L3):
\begin{center}
\begin{tabular}{@{}ll@{}}
L1 & learn on the provisional outcome, score on it \\
L2 & learn on the provisional outcome, score on the settled one \\
L3 & learn on the settled outcome, score on it, released only after its publication delay $D$ \\
\end{tabular}
\end{center}
L2 uses the same predictions as L1. It is what an operator faces when learning from the provisional outcome, as archived in the package, while the settled figures arrive later. Under L3, the maturation rule is extended by $D$, so that an outcome for origin $o$ is released only at $o + H + D$.

Under L3, the look-back is still drawn from the provisional outcome, because at any origin the operator has that version up to the present but has the settled one only up to $D$ days earlier. We run $D=0$, the hypothetical case in which the settled outcome arrives as fast as the provisional one, and $D=30$ days. At the documented upper bound of 90 days, the held-out layout no longer fits in four years of data. The warm slice alone spans 291 origins against 297 available, which leaves almost nothing for the early-stopping tail and the fit region together. We report that setting as infeasible under the second applicability condition and do not run it on a shortened slice.

\subsection{Correcting the load forecast}
\label{sec:opsd-l1}

Table~\ref{tab:opsd-l1} gives L1, with all three single correctors of Section~\ref{sec:setup} beside the two gates. The gated combination lowers mean MSE in all seven zones, and its worst cell over the 35 zone-run cells is $+0.37$\%. No single corrector is safe on this forecaster by the worst-cell criterion. The hardest cells are on Denmark. There the held-out corrector alone costs $+140.2$\% and the online corrector alone $+117.4$\%, while the three-expert gate keeps the same cell at $+0.37$. Denmark has the most accurate TSO forecast of the seven, and there the correctors add only variance. Averaged over runs, the two correctors cost between 20 and 56\% on Croatia, Belgium and Italy.

The static corrector deteriorates on no benchmark pair, but here, averaged over runs, it deteriorates three zones, none by more than one percent. Its worst cell is $+3.45$\%, on Denmark. That is small beside what the other two single correctors cost on the same cell. On Germany and Italy, the static corrector matches the three-expert gate to within a third of a point. On Hungary, it stops at $-41.1$\%, while the gate moves its weight to the online expert and reaches $-57.3$. The two correctors of the library are complementary on this data as on the benchmarks, and the gate improves both the zone where the static corrector deteriorates and the zone where it stops short.

The two gates differ in how they behave where correction does not help. On Denmark, Croatia, Italy and Belgium, the two-expert gate sits within a fiftieth of a percent of the TSO forecast. It places essentially all weight on the forecast and applies no correction. On Hungary, the same gate matches the online corrector to every reported digit, because it places all its weight on that corrector instead. The three-expert gate does neither. It keeps 0.019 of its weight on the static expert in Hungary. On Denmark, it averages $-0.5$\%, although that zone also holds the single worst cell of the load study. It improves all four zones where the two-expert gate applies no correction, including $-10.1$\% on Italy.

Larger base model error is associated with larger improvement across the seven zones, from $-0.09$\% on Belgium to $-57.3$\% on Hungary. Across zones, the weight that the gate keeps on the TSO forecast falls as the forecast's own MSE over the whole window rises (Spearman $-0.75$, $n = 7$, exact two-sided $p = 0.066$). The relation does not hold at every step of the ordering. Denmark has the most accurate forecast, and the gate keeps 0.637 of its weight on it. Italy has the second most accurate forecast but keeps under a tenth, because correction reduces the error so much there. Hungary, with the weakest forecast, is the only zone where the gate moves nearly all weight to the online expert. Even there it keeps a little weight on the other experts, unlike the near-total selection on the statistical base models of Table~S12.

\begin{figure}[pos=!ht]
\centering
\includegraphics[width=0.85\columnwidth]{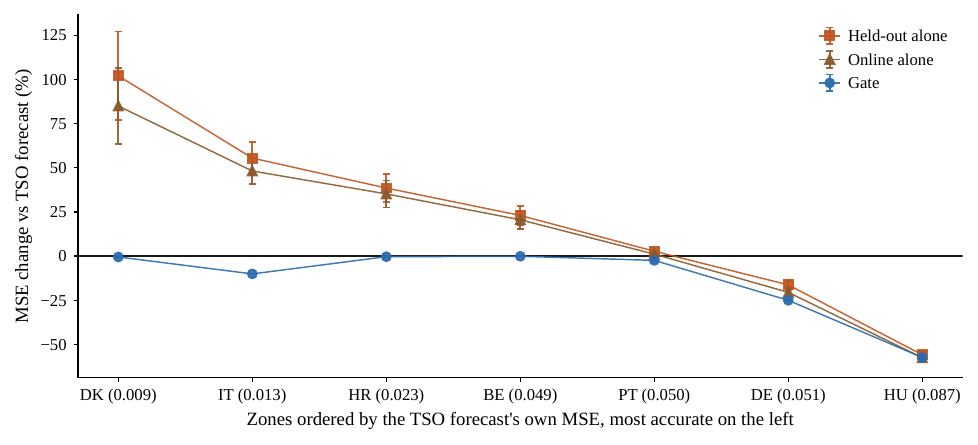}
\caption{Change against the TSO forecast by zone, five-run mean, ordered by that forecast's own MSE, in parentheses. Single correctors are harmful where the forecast is already accurate; the gate is the three-expert gate.}
\label{fig:zones}
\end{figure}

Fig.~\ref{fig:zones} orders the seven zones by the TSO forecast's own error, which spans a tenfold range. Along that order, the single correctors move steadily from doing harm to doing good, without a reversal. The held-out corrector runs from $+102.0$\% on Denmark to $-55.8$ on Hungary, and the online corrector spans a similar range over the same ordering. Their sign changes between Portugal and Germany, two zones whose TSO forecasts are almost equally accurate (Table~S18). An operator deciding whether to correct at all would therefore need to tell apart two zones that the accuracy figures barely separate. The gate does not follow that ordering. Where the single correctors are harmful, it stays within half a percent of the TSO forecast on Denmark, Croatia and Belgium, and it still finds $-10.1$\% on Italy. Where they gain, it takes as much as $-57.3$\%. Nothing in the layer measures the forecast's own quality, so the gate arrives at this separation from realized losses alone.

\paragraph{Sensitivity to the zone screen}
The quality screen removed the Netherlands because of a defect episode in its TSO forecast. To check whether the seven-zone result depends on that removal, we also run the Netherlands. The defect is one five-day episode entirely inside the training range, years before the first test origin, so no test origin's look-back reaches it. On the Netherlands, the combination improves the TSO forecast by 15.0\%, and it improves in every run. The mean over zones moves by less than a fifth of a point when the eighth zone is added, and the worst zone is unchanged at $-0.09$\%. Under the settled outcome, the two single correctors each cost about 20\% on this zone while the combination still gains 8.2\%.

\begin{table}[pos=!ht]
\caption{L1, learn and score on the provisional outcome. MSE change versus the TSO forecast in percent, mean $\pm$ standard deviation over five runs, all origins. Worst is the largest deterioration over zone-run cells.}
\label{tab:opsd-l1}
\centering
\footnotesize
\begin{tabular}{lr@{\,$\pm$\,}lr@{\,$\pm$\,}lr@{\,$\pm$\,}lr@{\,$\pm$\,}lr@{\,$\pm$\,}lr}
\toprule
Zone & \multicolumn{2}{c}{Static} & \multicolumn{2}{c}{Online} & \multicolumn{2}{c}{Held-out} & \multicolumn{2}{c}{Gate, $K{=}2$} & \multicolumn{2}{c}{Gate, $K{=}3$} & $w_0$  \\
\midrule
HU & $-41.08$ & $0.19$ & $-57.48$ & $2.02$ & $-55.84$ & $2.23$ & $-57.48$ & $2.02$ & $-57.34$ & $2.08$ & 0.000  \\
PT & $-2.84$ & $0.29$ & $+1.11$ & $2.03$ & $+2.76$ & $2.10$ & $-0.13$ & $0.24$ & $-2.42$ & $0.27$ & 0.217  \\
HR & $+0.09$ & $0.52$ & $+35.18$ & $7.64$ & $+38.50$ & $7.97$ & $+0.00$ & $0.00$ & $-0.33$ & $0.22$ & 0.494  \\
BE & $+0.52$ & $0.29$ & $+20.55$ & $5.03$ & $+23.01$ & $5.48$ & $-0.02$ & $0.04$ & $-0.09$ & $0.14$ & 0.549  \\
DK & $+1.00$ & $2.11$ & $+84.92$ & $21.44$ & $+101.96$ & $25.10$ & $+0.00$ & $0.00$ & $-0.48$ & $0.77$ & 0.637  \\
DE & $-25.10$ & $0.47$ & $-20.51$ & $2.57$ & $-16.28$ & $2.72$ & $-12.78$ & $4.38$ & $-25.00$ & $0.46$ & 0.003  \\
IT & $-9.76$ & $0.35$ & $+48.15$ & $7.45$ & $+55.45$ & $8.92$ & $-0.00$ & $0.00$ & $-10.06$ & $0.32$ & 0.088  \\
\midrule
Worst & \multicolumn{2}{r}{$+3.45$\hphantom{\,$\pm$\,0.19}} & \multicolumn{2}{r}{$+117.40$\hphantom{\,$\pm$\,21.44}} & \multicolumn{2}{r}{$+140.25$\hphantom{\,$\pm$\,25.10}} & \multicolumn{2}{r}{$+0.00$\hphantom{\,$\pm$\,2.02}} & \multicolumn{2}{r}{$+0.37$\hphantom{\,$\pm$\,2.08}} &   \\
\bottomrule
\end{tabular}
\end{table}

\paragraph{Classical weighting on the load data}
The intercept corrections of Section~\ref{sec:main-results} and the equal-weight average behave here in the opposite way to the benchmarks (Table~S8). The intercept correction, which improves no benchmark pair, improves six of seven zones here. Its mean of $-18.1$\% exceeds the gate's $-13.7$, and it reaches beyond $-70$\% on Hungary. On those zones, the TSO forecast carries a slowly moving bias, consistent with the highly autocorrelated TSO errors reported by \citet{kazmi2022tso}. A running mean removes it, while the learned correctors, fitted on a fixed training split, remove it only in part. On Denmark, where the forecast is already most accurate, the same correction costs $+23.4$\% and the exponentially weighted correction costs $+1{,}441$\%, the largest deterioration in the study. Equal weights deteriorate Denmark and Croatia by several percent.

Across the seven zones, the gate is the only method that improves every one, and its worst zone-run cell stays below half a percent. The library of Section~\ref{sec:experts} does not contain the intercept correction, the strongest single expert for this data by mean, so adding it is a natural extension. With the intercept correction as a fourth expert and everything else unchanged, the gate improves the load mean to $-20.9$\%, beyond the standalone correction. However, it also deteriorates Denmark by $+5.0$\% and moves the worst benchmark pair to $+0.47$\%. The gain comes with the loss of that downside control, and Section~\ref{sec:warmtrust} shows where the loss arises.

Section~\ref{sec:main-results} and this section point in opposite directions about single correctors. On the benchmarks, the static corrector is the safe choice and the intercept corrections cost double digits. On the load data, the intercept correction is strong and the static corrector deteriorates for the first time. Neither regime announces itself in advance. Among the methods run in both studies, the three-expert gate is the only one whose observed worst case stays small in both, without being told which regime it is in.

\subsection{Intervals on the load forecast}
\label{sec:opsd-intervals}

The adaptive tracker of Section~\ref{sec:conformal} wraps the gated forecast in each zone with the same $\gamma$ and $\alpha$ as the benchmarks, neither tuned for this data. Table~\ref{tab:opsd-intervals} reports it at the 90\% level on all origins, as in Table~\ref{tab:opsd-l1}. It is compared with static split calibration on the same forecast and with the same tracker on the uncorrected TSO forecast, and the tracker updates on matured origins only. Split calibration sets its width from the TSO forecast's held-out residuals, which no component of the layout has used. The tracker on the gated forecast has the lowest Winkler score in all seven zones; on Denmark it ties with the tracker on the TSO forecast at the printed precision. Against split calibration, the margin is 21\% on Portugal, 5\% on Hungary and under 5\% elsewhere. The adaptive intervals are 1.3\% wider on average, but narrower on Hungary and Italy, where split calibration over-covers. Against the tracker on the TSO forecast, the correction lowers the score in every zone, only slightly on Croatia, Belgium and Denmark, where the correction itself is small. It also narrows the intervals on average, most on Germany and Hungary. The point correction therefore improves the intervals here as it does on the benchmarks.

Coverage is within 0.04 of nominal in five zones, but Portugal under-covers at 0.79, an error of 0.11 that the mean over zones does not show. Portugal shows that the tracker's long-run guarantee is not reached within this window, which matters where a stated level must hold zone by zone. Fig.~S1 extends the comparison to the levels 0.2 to 0.9. The two adaptive methods stay within 0.04 of nominal on average at every level. Split calibration drifts: it over-covers at the low levels and under-covers on Portugal at every level.

Table~S15 scores the same three methods by the mean pinball loss over the nine deciles. The tracker on the gated forecast has the lowest score in every zone. Its margin over the tracker on the TSO forecast follows the point correction: it is largest on Hungary and Germany and within 0.0002 on Croatia, Belgium and Denmark. Its margin over split calibration is largest on Hungary and Portugal. Here calibration and the proper score agree.

\begin{table}[pos=!ht]
\caption{Intervals at the 90\% level on the TSO forecast, learn and score on the provisional outcome, five-run mean, all origins. Coverage closer to 0.90 is better; lower width and Winkler are better. Under the two gates the run-to-run standard deviation is at most 0.00383 in every cell except Hungary's Winkler scores, where it is 0.0128 under the adaptive gate and 0.0109 under split calibration; the TSO forecast and the split-calibrated width carry no run variation.}
\label{tab:opsd-intervals}
\centering
\footnotesize
\begin{tabular}{lrrrrrrrrr}
\toprule
& \multicolumn{3}{c}{Gate, adaptive} & \multicolumn{3}{c}{Gate, split} & \multicolumn{3}{c}{TSO, adaptive}  \\
\cmidrule(lr){2-4}\cmidrule(lr){5-7}\cmidrule(lr){8-10}
Zone & Cov & Width & Winkler & Cov & Width & Winkler & Cov & Width & Winkler  \\
\midrule
HU & 0.941 & 0.796 & 0.922 & 0.964 & 0.892 & 0.971 & 0.883 & 0.914 & 1.138  \\
PT & 0.792 & 0.784 & 1.449 & 0.680 & 0.642 & 1.841 & 0.789 & 0.791 & 1.467  \\
HR & 0.888 & 0.489 & 0.695 & 0.888 & 0.497 & 0.699 & 0.887 & 0.490 & 0.696  \\
BE & 0.896 & 0.613 & 0.789 & 0.915 & 0.653 & 0.795 & 0.896 & 0.613 & 0.790  \\
DK & 0.888 & 0.229 & 0.367 & 0.853 & 0.205 & 0.383 & 0.888 & 0.229 & 0.367  \\
DE & 0.862 & 0.695 & 0.978 & 0.823 & 0.636 & 1.014 & 0.835 & 0.760 & 1.126  \\
IT & 0.917 & 0.350 & 0.445 & 0.935 & 0.380 & 0.456 & 0.909 & 0.361 & 0.464  \\
\midrule
Mean $|\mathrm{cov}-0.90|$ & 0.033 & & & 0.067 & & & 0.033 & &  \\
Mean & & 0.565 & 0.806 & & 0.558 & 0.880 & & 0.594 & 0.864  \\
\bottomrule
\end{tabular}
\end{table}

\subsection{Learning and scoring on different outcome versions}
\label{sec:opsd-labels}

Table~\ref{tab:opsd-labels} crosses the outcome versions. Scored on the settled outcome, the same L1 predictions improve only four of seven zones. Italy, where the revision is largest (Table~\ref{tab:opsd-labels}), turns from $-10.1$ to $+6.2$\%. The reversal appears in all five runs, and the Diebold--Mariano test rejects equal accuracy in both directions on each. Every expert's forecasts are unchanged while the outcome moves, so relative to the settled outcome all of them are biased in the same direction. A combination confined to the simplex cannot remove a bias that its experts share. The static corrector, an expert rather than a combination, moves on Italy from $-9.8$ to $+6.9$\% under the same crossing (Table~\ref{tab:opsd-labels}). The bias is therefore present in the experts before any weight is applied.

Suppose the settled outcome differs from the provisional one by a common term $b$, so that $y^{(S)} = y^{(P)} + b$. Then every expert's error against the settled outcome is its error against the provisional outcome plus that same term, $e_k^{(S)} = e_k^{(P)} + b$. For any weights on the simplex,
\begin{equation}
\sum_k w_k\, e_k^{(S)} = \sum_k w_k\, e_k^{(P)} + b \sum_k w_k = \sum_k w_k\, e_k^{(P)} + b.
\label{eq:sharedbias}
\end{equation}
Whatever the weights are, the shared term passes through the combination unchanged. Removing an additive shared term calls for an intercept rather than a different weight sum: weights that sum to $s$ turn the term into $sb$, which vanishes only with the weights themselves. The fourth-expert experiment above is that intercept route, and it trades the shared term for the loss of downside control. A multiplicative revision behaves differently: under $y^{(S)} = r\,y^{(P)}$, weights that sum to $r$ absorb it, so leaving the simplex helps exactly when the revision scales the outcome. On the load data the revision is a drifting ratio, so Equation (\ref{eq:sharedbias}) is an idealization that the observed pattern approaches but does not match. \citet{radchenko2023similar} reach weights outside $[0, 1]$ by a different route, conditioning on information that makes the experts' biases differ, with the weights still summing to one.

Learning on the settled outcome at $D=0$ and at $D=30$ days restores the improvement in all seven zones. The size of the restoration rises with the size of the revision (Spearman $0.89$ at $D{=}0$ and $0.82$ at $D{=}30$, $n = 7$; exact two-sided $p = 0.012$ at $D{=}0$). Hungary, where the two versions barely differ, gains nothing from the change of outcome, while Italy returns to $-50.2$\% at $D = 30$. The delay itself costs at most 2.9 points on four zones, and 6 to 18 points on Hungary, Germany and Italy, where the combination gains most. It never reverses a sign. Germany at $D = 30$ is the one zone whose allocation is not stable across runs. Each run settles on either the online expert or the static one, with none in between. The run dispersion around its mean gain of $-43.7$\% is the widest in the table.

\begin{table}[pos=!ht]
\caption{Scored on the settled outcome, mean $\pm$ standard deviation over five runs. Revision is the mean absolute relative difference between the two outcome versions. Every column from L2 onward is the MSE change versus the TSO forecast in percent: L2 and the two L3 columns are the gated combination under the corresponding outcome protocol, Scalar is a deterministic post-hoc per-hour-of-day ratio with no run replication, and Static, L2 is the static corrector.}
\label{tab:opsd-labels}
\centering
\footnotesize
\begin{tabular}{lrr@{\,$\pm$\,}lr@{\,$\pm$\,}lr@{\,$\pm$\,}lrr@{\,$\pm$\,}l}
\toprule
Zone & Revision (\%) & \multicolumn{2}{c}{L2} & \multicolumn{2}{c}{L3, $D{=}0$} & \multicolumn{2}{c}{L3, $D{=}30$} & Scalar & \multicolumn{2}{c}{Static, L2}  \\
\midrule
HU & 0.59 & $-53.23$ & $1.84$ & $-52.74$ & $1.91$ & $-46.85$ & $0.86$ & $+33.36$ & $-41.53$ & $0.20$  \\
PT & 2.22 & $-0.10$ & $0.21$ & $-9.30$ & $0.14$ & $-9.24$ & $0.16$ & $-12.69$ & $-0.22$ & $0.24$  \\
HR & 2.60 & $-0.63$ & $1.17$ & $-15.59$ & $0.45$ & $-15.67$ & $0.33$ & $-4.35$ & $-0.82$ & $2.47$  \\
BE & 3.18 & $+0.16$ & $0.40$ & $-9.37$ & $4.07$ & $-6.50$ & $1.01$ & $-9.03$ & $+0.44$ & $0.82$  \\
DK & 3.87 & $+0.43$ & $0.49$ & $-12.18$ & $0.69$ & $-11.16$ & $0.49$ & $+66.43$ & $+1.78$ & $1.03$  \\
DE & 5.93 & $-25.04$ & $0.62$ & $-53.05$ & $3.47$ & $-43.71$ & $4.84$ & $-38.71$ & $-25.13$ & $0.62$  \\
IT & 9.40 & $+6.20$ & $0.19$ & $-68.44$ & $2.27$ & $-50.25$ & $4.08$ & $-81.79$ & $+6.88$ & $0.26$  \\
\bottomrule
\end{tabular}
\end{table}

\begin{figure}[pos=!ht]
\centering
\includegraphics[width=0.9\columnwidth]{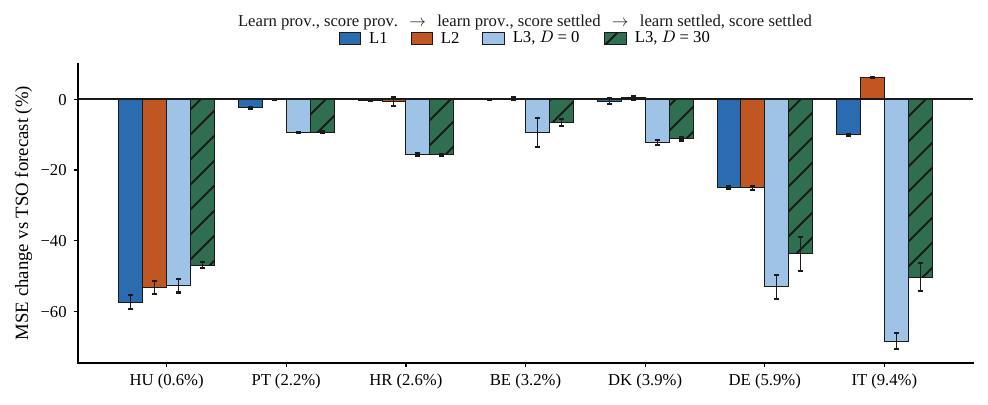}
\caption{Learning and scoring outcome versions crossed, seven zones, five-run mean. Scored on the settled outcome, four of the seven zones still improve; learning on the settled outcome restores all seven.}
\label{fig:labels}
\end{figure}

Fig.~\ref{fig:labels} places the four protocols side by side, with the zones ordered by the size of their revision. The zero line is what matters in this figure. Under L1 and under both L3 delays, every bar sits below it. Under L2, three bars cross it, for Belgium, Denmark and Italy, and the improvement on Portugal and Croatia falls under 1\%. The same effect appears as the gap between the L2 and the L3 bars. The gap grows across the panel, from a fraction of a point on the left to tens of points on the right, although not strictly zone by zone. Germany breaks that ordering but not the effect. Its revision is the second largest, yet L2 still improves it by $25.0$\%. Revision size therefore marks where the risk lies but does not fix what it costs.

\paragraph{Comparison with a level correction}
When two outcome versions differ by a roughly constant factor, a corrector can obtain most of its gain by learning that factor. The last column of Table~\ref{tab:opsd-labels} reports a baseline that multiplies the TSO forecast by a per-hour-of-day ratio of the settled to the provisional outcome, fitted on the training split. Where the ratio is stable, that baseline is enough. In Italy it reaches $-81.8$\%, beyond the gated combination at either delay, and Germany shows a smaller effect of the same kind. Where the ratio is small or drifts, the baseline is harmful and the combination is not. In Hungary and Denmark, the fixed correction makes the forecast worse by 33 and 66\%, while the combination under L3 at $D{=}30$ reaches $-47$ and $-11$\%.

Denmark shows why. The training-split ratio exceeds the test-period ratio by three points, in line with the fluctuating deviations between the two sources that \citet{hirth2018entsoe} find for Denmark, so the fixed correction over-corrects a forecast whose error is the smallest of the seven. A 3\% over-correction costs 66\% of MSE. The gate weights its own experts by their realized losses, so an expert that over-corrects has its weight reduced. Across the seven zones, the revision size marks where scoring on the settled outcome is risky, but it does not fix the size of what the combination recovers.

The method cannot repair a bias shared between the learning outcome and the scoring outcome. When the two outcomes are aligned, the combination adds a gain that a fixed level correction does not consistently match. The scalar baseline is the better method on Italy, where the revision is close to a stable ratio, but it costs 66 and 33\% on Denmark and Hungary, where the revision is not.

\paragraph{Pre-specified choices}
Zone selection, the protocol, the outcome-version crossing and the first two zone screens of Section~\ref{sec:opsd-data} were fixed before the corresponding runs, and Section~\ref{sec:opsd-data} states when the third screen was added. The hourly-ratio baseline and the train-versus-test ratio diagnostic were added after the outcome-version results were seen, and the caption of Table~\ref{tab:opsd-labels} marks the baseline as post-hoc. Gate weights in Table~\ref{tab:opsd-l1} are time averages over the matured test stream. All cells use five runs, and the Diebold--Mariano statistics are computed on each run separately.

\section{Discussion}
\label{sec:discussion}

\subsection{Two failed extensions}
\label{sec:quasistatic}

The gate's downside control is not unconditional. Two extensions that failed locate the conditions. The first adds an expert that keeps learning, and the second shortens the region on which the warm start is fitted. The first extension adds a fourth expert, the calibration module of PETSA \citep{petsa2025} restricted to our matured-outcome rule, and this breaks the downside control. The worst pair on DLinear moves from $-0.30$ to $+1.22$\%. On ETTh2, the four-expert combination is worse than every one of its own experts, including the base model.

Convexity gives the combination (\ref{eq:combination}) a pointwise bound: at every origin, it is no worse than the worst expert forecast at that origin. With time-varying weights, however, this does not bound the cumulative error against every fixed expert, so the combination can in principle be worse than all four experts. The violation traces back to where the weights come from. The weights are computed from losses that matured $H$ origins earlier, so an expert whose parameters move appreciably over that interval is scored on a version of itself that no longer exists.

The learning rate confirms the mechanism directly (Table~S21, panel A). Slowing the fourth expert tenfold makes it both better on its own and harmless inside the gate. Slowing it a further tenfold keeps the combination's downside small. The two slowed rates were chosen after test performance had been observed. The table is therefore a diagnosis of the mechanism, not a configuration we recommend.

The resulting condition is that an expert must change little over the maturation delay. We call such an expert quasi-static. For the one expert that crossed this boundary, panel A of Table~S21 places it between the published learning rate and a tenth of it. In practice, we use the update cadence of Section~\ref{sec:method} as the working definition: one optimizer step per 64 matured origins on the benchmarks and one per 8 on the load streams. Cadence is our proxy, not a general measure of how far an expert moves. Our three experts satisfy the condition. The base model and the static corrector do not move at all, and the online corrector updates slowly relative to $H$. The condition is about speed, not quality. On Electricity, the added expert under our outcome protocol is roughly 1\% better than the base model but far weaker than the other experts. The four-expert gate still delivers $-2.47$\% there. Under its native protocol, the same module makes the base model 2.9\% worse. Any mechanism that tracks experts through delayed feedback inherits this constraint. Delay-aware weighting is therefore the natural direction for admitting fast-moving experts.

\label{sec:warmtrust}

A short fit region leaves an allocation that does not transfer to the test stream, and the maturation delay makes any recovery late. The warm start replays the gate update on the warm slice so that the test stream does not carry the burn-in. The slice and the early-stopping tail are reserved first, and a pair is admitted whenever any fit region remains after them. A short held-out split therefore shrinks the fit region, and with it the reliability of the allocation that the warm start returns.

One pair shows both the short fit region and the unreliable allocation it leaves. Exchange with a DLinear base model leaves 303 held-out origins for the fit region at horizon 96 and only 120 at horizon 192. The latter is by far the smallest fit region among the 19 pair-horizon configurations run at the longer horizons. At horizon 336, the slice alone would require 536 origins against a held-out split of 425, so the pair is refused outright. A corrector fitted on that region alone is a poor forecaster at either horizon. The held-out corrector, fitted on nothing else, costs 90.1\% against the base model at horizon 192 (Table~S20) and 83.7\% at horizon 96 (Table~\ref{tab:main}). At horizon 96, the combination still improves. The two horizons differ in the weight that the warm start assigns to the online expert: zero in every run at horizon 96, and above 0.96 in four of five runs at horizon 192. On the slice, the online corrector, frozen in its fit-region state, has a low loss. That allocation is wrong on the test stream, and the combination ends 23.0\% worse than the base model it started from.

The gate is not too slow to recover: it updates once per matured origin and drives the weight on the losing expert to zero within the first quarter of the stream. The maturation rule makes the recovery late. No loss is observable until a forecast's horizon has elapsed, so the first $H$ origins are all issued under the warm-start allocation. In the traced run, those origins account for most of the excess error over the whole stream. Across the five runs, the initial weight on the losing expert tracks the outcome closely. In the one run where the warm start did not concentrate on that expert, the pair does not deteriorate. The warm start is only as useful as the fit region behind it, and a layout that protects the slice at the fit region's expense turns it into a cost.

The fourth expert of Section~\ref{sec:opsd} fails in the same way on a full fit region. Panel B of Table~S21 sets the four-expert gate beside the three-expert gate and the intercept correction alone. The added expert is quasi-static on the scale of the delay, so the condition of Section~\ref{sec:quasistatic} is met, and the gate does capture what the expert offers. The load mean under L1 exceeds that of the standalone correction, and under L3 with no delay every zone improves at a mean of $-43.4$\%.

With the fourth expert admitted, the downside control fails on Denmark. There the intercept correction alone costs $+23.4$\%, but it has a low loss on the warm slice, so the warm start assigns it 0.60 of the initial weight. The gate drives that weight to 0.004 on average over the stream. However, the origins issued before the first losses mature use the slice's allocation, and they cost the zone $+5.0$\%. On the benchmarks, the same mechanism moves the worst pair from $+0.15$ to $+0.47$\%. That pair is ETTh2 with DLinear, which the three-expert gate improves; the two-expert gate's figure of the same size in Section~\ref{sec:ranks} is on a different pair. Under the delayed outcome at $D = 30$, it costs Belgium $+16.0$\%, while the three-expert gate gains 6.5\% there. The gate is built for an expert that is strong on most zones and harmful on a few, and it does contain the harm once the stream is running. It cannot yet reject the slice's allocation for that expert before the first test losses arrive. The three-expert library is therefore the reported configuration.

\subsection{Three applicability conditions}
\label{sec:conditions}

These are the empirical operating conditions, identified on the evaluated streams, under which the regret bound of Section~\ref{sec:gate} is a meaningful statement about the frozen base model. Each is stated with the experiment and the number that fix it. The bound itself holds without them; what they protect is the comparison, keeping the frozen base model a comparator worth bounding against over the delay.

\subsubsection{Expert stability}
Experts must be quasi-static on the timescale of the maturation delay. This condition comes from Section~\ref{sec:quasistatic}. An expert that moves between the origin at which the gate scores it and the origin at which its forecast is used is weighted on an earlier version of itself, so the gate reacts to stale performance.

\subsubsection{Stream sufficiency}
The held-out split must host the layout and leave a fit region on which the warm start is reliable. The test stream must also be long enough for the gate to learn from matured losses. A feasible layout is necessary but not sufficient. This condition comes from Section~\ref{sec:warmtrust}. The held-out data must contain a warm slice of $H + 200$ matured origins, disjoint from the corrector's fit region and from the early-stopping tail. The fit region that remains must contain at least $H$ matured origins, so that the slice's allocation for the online corrector transfers to the test stream. The test stream must contain enough matured origins for the gate's allocation to overtake a fixed corrector.

The weekly influenza-like-illness series of the long-horizon benchmark collection has 966 rows and fails the first requirement in all 12 configurations we tried. Its held-out split supplies at most $195 - H$ origins against a requirement of 201. At the most generous split and the shortest horizon tested, the series would need 1{,}358 rows. Shortening the slice rather than skipping it does not help. A short slice evaluates the online corrector before it has adapted. The resulting warm start is no better than a uniform one in any configuration, and worse by up to 1.75 points. On the 98 to 146 matured test origins that such a series provides, the plain held-out corrector has lower error than the gated combination in all 50 matched comparisons we ran, by up to 5.46 points.

The same limit binds on the load data at the documented ninety-day publication delay of the settled outcome, where the warm slice alone takes 291 of the 297 held-out origins (Section~\ref{sec:opsd}). Panel A of Table~S17 gives the full layout per zone and delay. At the delays that do run, 34 to 66 daily origins remain for the fit region after the slice and the 30-origin tail. These are the smallest fit regions in the study. The zone-level outcomes of Section~\ref{sec:opsd} verify that the slice's allocation transfers there. Every other admitted configuration in the study clears that floor. The one that does not is Exchange with DLinear at horizon 192, where the layout is admitted with a fit region of 120 origins against a horizon of 192, and the resulting allocation costs 23.0\%. Sections~S10 and~S11 give the feasibility map and the longer-horizon results.

In our setting, the slice allocation failed to transfer at 120 fit origins and held at 303 (Table~S17, panel B). With 98 to 146 matured test origins, the gate cannot learn the allocation at all (Section~S10). These values locate the failure boundary in our setting; of them, only the fit-region floor is carried forward as a requirement.

\subsubsection{Outcome alignment}
The learning outcome and the scoring outcome must be the same version. This condition comes from Section~\ref{sec:opsd}. When the stream runs on a provisional outcome and is scored on a settled one, every expert's error shifts by the same revision, so all of them are biased in the same direction relative to the score. By (\ref{eq:sharedbias}), a combination confined to the simplex cannot remove a bias that its experts share, and the optimal weight in that case lies outside $[0, 1]$ \citep{radchenko2023similar}. Where the two versions nearly agree, the improvement survives the crossing, as on Hungary in Table~\ref{tab:opsd-labels}. The regret bound (\ref{eq:regret}) concerns the outcome the gate observes. It carries over to a different outcome only to the extent that the two agree.

\subsection{Limitations}
\label{sec:limitations}

The gate forms weighted averages, so on pairs where one expert is clearly best it still keeps some weight elsewhere. For this reason, the two-expert gate stays behind the best single expert on three foundation-model pairs, each time by less than two thirds of a percentage point. Sharper allocation rules are natural candidates: fixed-share updates \citep{herbster1998} and second-order rules such as Bernstein online aggregation \citep{wintenberger2017boa}. Replayed over the same experts, the same warm-slice replay and the same maturation rule, neither separates from the Hedge update. On the time-averaged-weight basis of Table~S22, fixed-share at $\alpha = 1/T_m$ lands 0.05 percentage points from the Hedge gate and Bernstein aggregation 0.15, both inside the 0.52-point mean gap that the pricing basis itself carries, while fixed-share at $\alpha = 0.01$ trails by 1.41 points. The worst-pair margin thins to $-0.00\%$ under Bernstein aggregation where the Hedge gate keeps $-0.17\%$, and on the load streams the same rule matches the gate under L1 but falls behind under the thirty-day settled-outcome protocol, where its self-tuned rates concentrate on recent evidence that the maturation delay has made stale. Keeping the update standard leaves every degree of freedom, the maturation rule, the warm slice, the trust region and the update cadence, in the pre-specified protocol, where each is inspectable rather than absorbed into the optimizer. Sleeping experts and per-channel weights remain untried.

The maturation rule discards partially observed horizons. As the PETSA comparison shows, this costs accuracy where a horizon can legitimately be scored before it completes. A partial-outcome variant is compatible with the architecture and is left open.

The interval layer does not model conditional error scale. It improves calibration under shift, through the adaptive tracker on nearly every pair and through the correction on most pairs. The correction narrows the intervals only where it lowers the error. A method that improved sharpness on its own would have to model that scale. On the load data, one zone under-covers by 0.11 at the 90\% level within a four-year window, so the long-run guarantee should not be read as a per-zone one.

Finally, the evaluation covers standard multivariate benchmarks and one operational dataset with one revision mechanism, a re-metered load series. The provisional series enters as archived in the package and may include later platform-side corrections, a possibility that \citet{kazmi2022tso} also note for data downloaded from the platform. The crossing therefore brackets the true first-release vintage rather than matching it exactly. The deployment argument is strongest for streams such as epidemiological surveillance and commodity markets. These streams have retrospective and repeated revisions and regime changes that neither the benchmarks nor the load data show, and testing there is the most informative next step.

\section{Conclusions}
\label{sec:conclusion}

We treat the improvement of a forecaster that cannot be retrained as a deployment problem, with the objective of limiting deterioration relative to the forecast it starts from. The residuals of frozen base models on standard benchmarks are dependent over time on every dataset and run. Yet almost nothing in them can be recovered from a short summary of a single input window. Optimizing that summary against each sample's own outcome measures, at a stated budget, the most a controller could gain through it, without testing controllers one at a time. This result suggests a different goal: not a more accurate single corrector, but a rule that decides from observed performance how much of each corrector to use.

Our method holds the base model frozen and weights it against a static trust-region corrector and an online corrector, under weights that are non-negative and sum to one. The weights are updated multiplicatively from losses that become available only after the horizon has elapsed. They are warm-started by replaying the same update on a disjoint slice of the held-out data.

At the main horizon, across the 28 benchmark pairs, the worst deterioration is 0.15\% and the gains reach 11.5\%. Two of the four frozen base models are zero-shot foundation models the correctors were not designed against, and the downside stays small on them with nothing adjusted. On day-ahead load forecasts published by European transmission system operators, whose parameters are not available to users, the combination improves every bidding zone on the outcome it learns from. Each corrector run alone is harmful on at least one zone. On the zone whose forecast is already the most accurate, the held-out and online correctors cost 102\% and 85\%. The same tracker that calibrates the benchmark forecasts gives intervals on the load forecasts that score better than split calibration in every zone. The limited downside comes from the allocation among the experts, not from any one of them. Where the base model is weak enough, the same rule moves nearly all the weight onto a corrector, so one rule covers both adjusting a forecast and replacing it.

The load data carries its target twice: a provisional outcome published within hours and a settled one published months later. The two versions separate the outcome a method learns from and the outcome it is judged on. Scored on the settled outcome, the same predictions improve only where the two versions nearly agree. Weights confined to the simplex cannot remove a bias that every expert shares. Learning on the settled outcome at delays of zero and thirty days restores the improvement in every zone.

Three conditions limit where these results hold: expert stability, stream sufficiency and outcome alignment. Each comes from an experiment that failed, and each points to an extension. Weighting that accounts for the delay in observing each expert's losses would relax expert stability. Adding the fit-region floor of Section~\ref{sec:conditions} to the layout guard would enforce stream sufficiency before any outcome is seen; the layout check used here requires only that the region be non-empty. Outcome alignment could be addressed by estimating the shared term directly, with an intercept for an additive revision or a rescaled weight sum for a multiplicative one, at the price of the convexity that the combination relies on.

The load data adds a fourth extension. An intercept correction outside the library improves six of the seven zones and has a better load mean than the gate, but it is harmful on one zone. Admitting it as an expert raises the load mean beyond that of the correction itself, but gives up the downside control on the zone where it is harmful. A warm start that gives little weight to an expert whose slice performance does not carry over to the stream would let such an expert in.

Denmark has the most accurate TSO forecast of the seven zones, and the gate leaves it nearly untouched. Hungary has the weakest, and there the gate moves almost all of its weight onto a corrector and more than halves the error. How much to adapt is a property of the forecast being adapted, and no fixed amount chosen before deployment is right for both zones. A frozen forecast therefore need not be replaced to benefit from adaptation. Two provisos apply: outcomes must mature before they influence the allocation, and the layer must learn from the outcome it will be judged on. Under those conditions, the layer can be attached to a forecast nobody can retrain, and its downside is measured against the forecast it started from, not against the average quality of its corrections. This is a statement about deployment, not only about benchmarks, because adaptation is evaluated with the information an operator holds at decision time, not with ground truth that arrives months later. Before deployment, an operator therefore does not need to decide how much to adapt, only whether the layer applies. That means checking three things: whether each corrector changes little over the maturation delay, whether the held-out data holds the warm slice with a fit region to spare, and whether the outcome used for learning is the one the forecast will be judged on.

\section*{Data and code availability}

The load data are the public ENTSO-E package cited in Section~\ref{sec:opsd}, and the benchmark series are the public archives cited in Section~\ref{sec:setup}. The code, the stored run set and the result files behind every table will be made available in a permanent public repository on publication, with a \texttt{README} that lists the entry point for each table and figure. The supplementary material holds Sections S1 to S14, Tables S1 to S23 and Fig.~S1, which the main text refers to by those numbers.

\section*{Declaration of generative AI and AI-assisted technologies in the manuscript preparation process}

During the preparation of this work the authors used Claude to refine the wording of the manuscript. The authors then reviewed and edited the text as needed and take full responsibility for the content of the published article.

\bibliographystyle{elsarticle-harv}
\bibliography{refs}
\end{document}


\begin{center}
{\large\bfseries Supplementary Material}\\[0.4em]
{\large Downside-Controlled Online Forecast Combination under Delayed and Revised Outcomes}
\end{center}

\vspace{0.8em}

\noindent Table~S1 defines the terms used throughout, and the sections follow the order of the main text. Section, table and figure numbers carry the prefix S, and references to sections, tables and figures without that prefix are to the main text. Where a table prints mean $\pm$ standard deviation over the five runs, the standard deviation carries one more decimal than the mean, or as many decimals as its first significant digit needs where it would otherwise print as zero.

\vspace{0.8em}

\section{Terms, data regions, datasets and settings}

\label{app:data-s}

\paragraph{Terms and data regions}

The term \emph{expert} of Table~\ref{tab:terms} comes from prediction with expert advice; the forecast combination literature calls the same object an individual or component forecast.

\begin{table}[!ht]
\footnotesize
\caption{Terms used throughout, with the symbol or the count each one names.}
\label{tab:terms}
\centering
\footnotesize
\begin{tabular}{@{}>{\raggedright\arraybackslash}p{0.23\textwidth} p{\dimexpr0.77\textwidth-2\tabcolsep\relax}@{}}
\toprule
Term & Meaning \\
\midrule
Frozen & A forecaster its user cannot retrain; its parameters are never updated in this study \\
Base model & The frozen forecaster the correction layer wraps, written $E_0$; on the load data it is the forecast of the transmission system operator (TSO) \\
Corrector & A module that alters the base model's forecast: the static corrector $E_1$ (a trust-region adapter fitted once on the training split), the online corrector $E_2$ (updated from matured errors during the stream) and, outside the library, the held-out corrector and the intercept corrections (running means of matured errors, plain or exponentially weighted) \\
Expert & A forecast the gate holds weight on: the base model $E_0$, the static corrector $E_1$ and the online corrector $E_2$; called a corrector ($E_1$, $E_2$) when its mechanism rather than its weight is discussed \\
Gate, Gate $K{=}n$ & The rule that sets the weight on each expert from the losses that expert has already accrued; $K$ is the number of experts, $K{=}3$ the reported configuration \\
Combination, gated combination & The single forecast the gate's weights produce; the row labelled Gate in every table \\
Correction layer, the layer & The whole proposed method wrapping the frozen base model: the experts, the gate that weights them, and the adaptive tracker's interval layer over the combination \\
Method & Any row a table compares: a single corrector, the base model, a fixed-weight rule or a gate \\
Pair & One dataset with one base model; seven datasets and four base models give 28 pairs \\
Zone, bidding zone & One bidding zone of the load data, with its TSO forecast as the frozen base model; the load study's unit in place of a pair \\
TSO, TSO forecast & Transmission system operator, the company that runs a bidding zone's high-voltage grid; its published day-ahead load forecast is the frozen base model of the load study and the quantity every load table reports a change against \\
Run & One random initialization of every trained component; each pair is run five times \\
Cell & The unit every worst-case statement quantifies over: one pair at one run (a pair-run cell) or one zone at one run (a zone-run cell); 28 pairs at five runs give 140 pair-run cells, and seven zones 35 zone-run cells \\
Worst pair, worst zone, worst cell & The largest MSE change over pairs, over zones, or over cells; worst cell is the strictest of the three \\
Origin & The last step of an input window, the point a forecast is issued from \\
Channel & One series of the multivariate target; $C$ channels are forecast jointly at every origin \\
Matured & An outcome fully observed: released $H$ steps after its origin, or $H+D$ steps under a publication delay \\
Maturation delay & The $H$ origins between issuing a forecast and observing its loss; every gate update is delayed by it \\
Publication delay $D$ & The further delay, in days, before the settled load is published; the load study runs $D$ of 0 and 30 days; distinct from the maturation delay \\
Held-out split & The data between training and test, partitioned into the three disjoint regions below \\
Fit region & The part of the held-out split the correctors are fitted on \\
Warm slice & The part the gate replays its update on, to set the weights the test stream opens with; with the fit region and the tail it forms the warm-start layout \\
Early-stopping tail & The part the correctors' stopping epoch is chosen on \\
Test stream & The test origins in chronological order; the gate updates on it and nothing else is fitted on it \\
Burn-in & The stream segment before the gate's weights settle; the warm start moves its excess loss off the test stream \\
Adaptive tracker & The adaptive conformal interval layer over the combined forecast; split calibration is the static split-conformal comparator on the same forecast \\
Provisional, settled & The two outcome versions of the load: the provisional outcome is published within hours on the ENTSO-E Transparency Platform, the settled outcome months later in ENTSO-E Power Statistics \\
L1, L2, L3 & The three load protocols: learn and score on the provisional outcome; learn on the provisional and score on the settled; learn and score on the settled with the publication delay $D$ added to the maturation rule \\
\bottomrule
\end{tabular}
\end{table}

\paragraph{Datasets and splits}

The regions of the held-out split and what each decides are stated in Table~1 of the main text; Table~\ref{tab:datasets} gives the series lengths and split sizes after windowing.

\paragraph{Decomposition parameters}
The residual audit and the ceiling test read their trend kernel and seasonal period from one configuration file. The online corrector's decomposition features read a separate per-dataset configuration, whose Weather entry is a period of 7 with a kernel of 48. ETTh1, ETTh2 and Electricity use a period of 24 with a kernel of 25, ETTm1 and ETTm2 a period of 96 with a kernel of 97, and Weather a period of 144 with a kernel of 145. The kernel and the period are capped at the length being decomposed. The audit decomposes the input window at $L{=}384$. The ceiling test decomposes the forecast at $H{=}96$, where a period of 96 or more leaves one point per phase class. On ETTm1, ETTm2 and Weather, the irregular branch therefore contributes a learned constant. Rerun at a sub-daily period of 24 under the protocol of Section~\ref{app:killtest}, the specified-budget ceiling stays under one percent on all three series (0.06, 0.01 and 0.82\%), so the reading of Section~3 is unchanged. At 2{,}000 steps, forty times the specified budget, Weather reaches 6.80\%, against 6.88\% at its native period.

\begin{table}[!ht]
\caption{Dataset statistics and split sizes as used, after windowing at $L{=}384$, $H{=}96$.}
\label{tab:datasets}
\centering
\footnotesize
\begin{tabular}{lrrrrl}
\toprule
Dataset & Channels & Train & Val & Test & Freq.  \\
\midrule
ETTh1/h2 & 7 & 8{,}640 & 2{,}880 & 2{,}880 & Hourly  \\
ETTm1/m2 & 7 & 34{,}560 & 11{,}520 & 11{,}520 & 15-min  \\
Weather & 21 & 36{,}887 & 5{,}270 & 10{,}539 & 10-min  \\
ECL & 321 & 18{,}412 & 2{,}632 & 5{,}260 & Hourly  \\
Exchange & 8 & 5{,}311 & 760 & 1{,}517 & Daily  \\
ILI & 7 & \multicolumn{3}{c}{966 rows, Section S10 of the supplement} & Weekly  \\
\bottomrule
\end{tabular}
\end{table}

ETT variants use the conventional 12, four and four month boundaries. The held-out split holds, in temporal order, a fit region for the correctors, a warm slice of $H + 200$ origins for the gate replay, and a 10\% early-stopping tail. Third-party baselines are aligned to the same boundaries, so their numbers here differ from their published ones.

\paragraph{Static corrector settings}

Table~\ref{tab:cleanroom} gives the settings of the static expert, Eq.~(3) of the main text, following the published description of the trust-region output adapter of \citet{deltaadapter2026}; rows that description does not fix are marked. Every setting is held fixed across every dataset, base model and horizon, except $\delta$, which takes one value on the ETT family and another elsewhere. Panel A of Table~\ref{tab:s-layout} halves and doubles it on the pair with the worst deterioration.

\begin{table}[!ht]
\caption{Settings of the static expert.}
\label{tab:cleanroom}
\centering
\footnotesize
\begin{tabular}{ll}
\toprule
Item & Setting \\
\midrule
Form & $y = \hat{y} + \delta \cdot A(\hat{y}, x)$ \\
Trust region & $\lVert A \rVert_\infty \leq 1$ via tanh \\
Radius $\delta$ & $0.01$ on ETT, $0.1$ otherwise \\
Optimizer & Adam, learning rate $10^{-4}$ \\
Network & Depth 2, width 128 \\
Fitted on & Training split, base model frozen \\
Early stopping & On the early-stopping tail \\
Epochs & 20 \\
Batch size & 64 \\
\bottomrule
\end{tabular}
\end{table}

\begin{table}[!ht]
\caption{The methods compared in Section~5 of the main text, what each is fitted on, and whether it is an expert in the gate's library. The documentation accompanying the code maps each name to its identifier in the result files.}
\label{tab:armmap}
\centering
\footnotesize
\begin{tabular}{llll}
\toprule
Method & Fitted on & Updates in the stream & In the library  \\
\midrule
Frozen base model & Not fitted here & No & $E_0$  \\
Static corrector & Training split & No & $E_1$  \\
Online corrector & Held-out fit region & Yes & $E_2$  \\
Held-out corrector & Held-out fit region & No & No  \\
Gate, $K{=}2$ & Carries $E_0$ and $E_2$ & Weights only &  \\
Gate, $K{=}3$ & Carries $E_0$, $E_1$ and $E_2$ & Weights only &  \\
\bottomrule
\end{tabular}
\end{table}

\section{Residual predictability audit}

\label{app:audit}

For each dataset, we train a DLinear base model \citep{Zeng2023DLinear} with input length 384 and horizon 96, freeze it, and examine its residuals $r_t = y_t - \hat{y}_t$ on the fit region of the held-out split, the region reserved for fitting correctors and disjoint from the test stream (Table~1 of the main text). We compute three diagnostics per channel; the Ljung--Box $p$ is reported as its maximum over channels and runs, the lag-one autocorrelation as its mean, and the ridge $R^2$ as the channel median averaged over five runs, because on Weather one heavy-tailed channel with folds of near-constant target dominates the channel mean; both summaries are in the result files. The first is a Ljung--Box test \citep{ljungbox1978} on the one-step-ahead residual series at lags 10, 24, and 48. The second is the five-fold cross-validated $R^2$ of a ridge regression with penalty $1$ on folds contiguous in time. The regression predicts the horizon-mean residual from per-sample statistics of a moving-average seasonal-trend decomposition of the input window: trend slope, seasonal amplitude, seasonal dominance, irregular standard deviation, and window mean and standard deviation. The decomposition takes a uniform moving average of the window as the trend, the per-phase mean of the detrended series as the seasonal part, and the arithmetic remainder as the irregular component. The third is the lag-one autocorrelation of the one-step residuals.

Table~\ref{tab:audit} reports the Ljung--Box test at lag 24; the conclusion is unchanged at the other lags. Negative $R^2$ indicates that the features predict the residual worse than its mean, the outcome on every dataset. Feature standardization is fitted on training folds only.

\begin{table}[!ht]
\caption{Residual predictability of a frozen DLinear base model on the held-out fit region. The Ljung--Box column is the maximum $p$ over channels and five runs; lag-one $\rho$ is the mean over channels and runs; ridge $R^2$ is the channel median averaged over runs. Ljung--Box rejects whiteness everywhere; ridge $R^2$ from per-sample statistics is negative everywhere.}
\label{tab:audit}
\centering
\footnotesize
\begin{tabular}{lrrr}
\toprule
Dataset & Ljung--Box $p$ (lag 24), max & Lag-1 $\rho$ & Ridge $R^2$  \\
\midrule
ETTh1 & $3.9\times10^{-5}$ & 0.089 & $-0.744$  \\
ETTh2 & $3.1\times10^{-4}$ & 0.108 & $-0.815$  \\
ETTm1 & $5.5\times10^{-12}$ & 0.225 & $-0.366$  \\
ETTm2 & $6.3\times10^{-6}$ & 0.109 & $-0.575$  \\
Weather & $8.2\times10^{-3}$ & 0.675 & $-0.149$  \\
\bottomrule
\end{tabular}
\end{table}

\section{Ceiling test: full results}

\label{app:killtest}

\paragraph{Corrector substrate}
The corrector is the sequence-level corrector of the online expert in Section~4 of the main text, extended with a conditioning interface. Each decomposition component of the base model forecast passes through its own $\mathrm{Linear}(H,H)$. A fourth branch maps an average-pooled projection of the look-back window through $\mathrm{Linear}(96,H)$, following adapters that condition on the input window \citep{deltaadapter2026,Liu2025PIR} and test-time adaptation \citep{tafas2025}. All correctors have between 28K and 37K parameters. A FiLM pathway \citep{Perez2018FiLM} inserts one hidden layer per component branch, width 48 for the sequence corrector and 64 for the pointwise one, whose modulation pair is generated from a conditioning vector $z \in \mathbb{R}^{8}$.

\paragraph{Measuring the ceiling}
The ceiling is measured by per-sample oracle probing. The corrector is trained with free per-sample embeddings $z_i$ optimized jointly with its weights \citep{glo2018} on the first two thirds of the held-out fit region, with its early-stopping tail inside that part; the last third is the probe region, which the corrector never sees. On the probe region the corrector is frozen, and $z$ is initialized at zero and optimized against each sample's outcome with Adam at learning rate 0.05. The resulting error is an outcome-informed benchmark for what any per-sample controller could deliver through an 8-dimensional bottleneck between the window and the corrector. A corrector that reads the look-back window directly, as the sequence substrate and the online expert do, lies outside this benchmark.

\paragraph{Result}
Table~\ref{tab:killtest-full} reports the outcome run by run. At the pre-specified budget of 50 steps no dataset exceeds 5\%. At the extended budget of 2{,}000 steps, $\lVert z \rVert$ reaches 80 to 224 and the rise is unstable across runs: ETTh1 exceeds 5\% on all five runs, ETTh2 and Weather on three of five, ETTm1 on two of five and ETTm2 on none.

\begin{table}[!ht]
\caption{Conditioning ceiling from per-sample oracle probing on the last third of the held-out fit region (737 origins on ETTh1 and ETTh2, 3{,}329 on ETTm1 and ETTm2, 1{,}454 on Weather), with the corrector fitted on the first two thirds (1{,}327, 5{,}992 and 2{,}616 training origins plus an early-stopping tail). Base MSE at $z{=}0$ on the probe region; ceiling at 50 steps, mean $\pm$ standard deviation over five runs. Extended-budget ceiling at 2{,}000 steps run by run with the five-run mean and standard deviation, and $\lVert z\rVert$ at 2{,}000 steps, mean $\pm$ standard deviation. The 5\% threshold was fixed before running.}
\label{tab:killtest-full}
\centering
\scriptsize
\setlength{\tabcolsep}{3.5pt}
\begin{tabular}{lrr@{\,$\pm$\,}lrrrrrrrr@{\,$\pm$\,}l}
\toprule
Dataset & $z{=}0$ MSE & \multicolumn{2}{c}{Ceiling, 50 steps} & \multicolumn{5}{c}{Per-run ceiling, 2{,}000 steps} & Mean & SD & \multicolumn{2}{c}{$\lVert z\rVert$, 2{,}000 steps}  \\
\midrule
ETTh1 & 0.6080 & \multicolumn{2}{r}{$3.76 \pm 1.03$\%\hphantom{\,$\pm$\,0.00}} & 26.55\% & 20.49\% & 19.50\% & 23.92\% & 20.76\% & 22.24\% & 2.92 & $79.939$ & $4.571$  \\
ETTh2 & 0.2998 & \multicolumn{2}{r}{$0.63 \pm 0.23$\%\hphantom{\,$\pm$\,0.00}} & 4.32\% & 6.58\% & 2.66\% & 6.48\% & 6.94\% & 5.40\% & 1.84 & $104.867$ & $12.760$  \\
ETTm1 & 0.3272 & \multicolumn{2}{r}{$0.14 \pm 0.19$\%\hphantom{\,$\pm$\,0.00}} & 0.95\% & 8.32\% & 5.93\% & 0.42\% & 0.07\% & 3.14\% & 3.75 & $223.934$ & $36.245$  \\
ETTm2 & 0.1055 & \multicolumn{2}{r}{$0.01 \pm 0.02$\%\hphantom{\,$\pm$\,0.00}} & 0.01\% & 0.01\% & 0.16\% & 1.17\% & 1.97\% & 0.66\% & 0.88 & $198.110$ & $53.313$  \\
Weather & 0.2715 & \multicolumn{2}{r}{$0.80 \pm 0.76$\%\hphantom{\,$\pm$\,0.00}} & 8.48\% & 8.99\% & 14.45\% & 1.56\% & 0.90\% & 6.88\% & 5.67 & $175.624$ & $77.876$  \\
\bottomrule
\end{tabular}
\end{table}

The oracle uses the probe region's outcomes and is a diagnostic, never a reported method.

\section{Reproducibility}

Every run reproduces bit for bit when repeated within the reference computing environment: random sources are fixed in advance, deterministic kernels are enabled, and error-covariance files are canonicalized before writing. Every number regenerates from the code, the input data and the run set alone, with no stored checkpoint. Third-party baselines are pinned by commit and never modified; the split alignment they require is applied from outside their code. The runs were executed on NVIDIA L40S devices, and the run records, the cache keys and the numerical tolerance across computing environments are documented with the code.

\section{Gate variants and classical weighting}

\label{app:classic}

Panel A of Table~\ref{tab:classic} summarizes the equal-weight average and the intercept corrections on the 28 benchmark pairs, Table~\ref{tab:opsd-classic} the seven load zones under L1, and panel B of Table~\ref{tab:quasistatic} the intercept correction admitted as a fourth expert. The equal-weight method averages the three experts with weight one third on every origin. The intercept correction adds the mean of the most recent 200 matured errors at the same lead. The exponentially weighted correction adds an exponentially weighted mean with a half-life of 100 origins. Both use whatever errors have matured, return the forecast unchanged before any error has matured, and have no fitted parameter or random source.

\begin{table}[!ht]
\caption{Classical weighting against the static corrector and the gate, MSE change versus the frozen base model in percent; pair and zone entries are five-run means. A, equal weights and intercept corrections on the 28 pairs, all origins, basis of Table~2 of the main text; worst cell is the maximum over the 140 pair-run cells. B, fixed and error-driven weight rules on the 28 benchmark pairs and the seven load zones, scored on matured origins, unlike Tables~2, 3, 5 and~6 of the main text.}
\label{tab:classic}
\centering
\footnotesize
\textit{A}\par\smallskip
\begin{tabular}{lrrrr}
\toprule
Method & Mean & Worst pair & Worst cell & Improved  \\
\midrule
Static corrector & $-2.24$ & $-0.09$ & $+0.08$ & 28/28  \\
Equal weights & $-1.54$ & $+9.06$ & $+11.74$ & 24/28  \\
Intercept correction & $+18.02$ & $+33.08$ & $+33.19$ & 0/28  \\
Exponentially weighted errors & $+13.07$ & $+22.18$ & $+28.26$ & 0/28  \\
Gate, $K{=}2$ & $-2.43$ & $+0.47$ & $+0.64$ & 21/28  \\
Gate, $K{=}3$ & $\mathbf{-3.99}$ & $+0.15$ & $+0.46$ & 25/28  \\
\bottomrule
\end{tabular}

\medskip
\textit{B}\par\smallskip
\begin{tabular}{lrrrrrr}
\toprule
 & \multicolumn{3}{c}{Benchmarks} & \multicolumn{3}{c}{Load}  \\
\cmidrule(lr){2-4}\cmidrule(lr){5-7}
Rule & Worst & Mean & Improving & Worst & Mean & Improving  \\
\midrule
Equal, all three & $+9.72$ & $-1.42$ & 24/28 & $+5.97$ & $-8.24$ & 4/7  \\
Equal, correctors only & $+17.06$ & $-1.13$ & 22/28 & $+17.44$ & $-7.30$ & 3/7  \\
Inverse recent error, $W=100$ & $+3.83$ & $-2.21$ & 24/28 & $+2.45$ & $-10.92$ & 4/7  \\
Inverse recent error, $W=500$ & $+3.13$ & $-2.30$ & 24/28 & $+3.48$ & $-10.85$ & 4/7  \\
Inverse recent error, all matured & $+2.78$ & $-2.35$ & 26/28 & $+3.48$ & $-10.85$ & 4/7  \\
Gate, $K{=}3$ & $+0.22$ & $-3.98$ & 23/28 & $-0.08$ & $-13.69$ & 7/7  \\
\midrule
In-sample optimum (oracle) & $-0.29$ & $-4.72$ & 28/28 & $-0.43$ & $-14.49$ & 7/7  \\
\bottomrule
\end{tabular}
\end{table}

\begin{table}[!ht]
\caption{Equal weights and intercept corrections on the TSO forecast under L1, basis of Table~6 of the main text. Mean $\pm$ standard deviation over five runs; the intercept corrections have no random source.}
\label{tab:opsd-classic}
\centering
\footnotesize
\begin{tabular}{lr@{\,$\pm$\,}lr@{\,$\pm$\,}lrrr@{\,$\pm$\,}l}
\toprule
Zone & \multicolumn{2}{c}{Static} & \multicolumn{2}{c}{Equal} & Intercept & EW errors & \multicolumn{2}{c}{Gate, $K{=}3$}  \\
\midrule
HU & $-41.08$ & $0.19$ & $-42.77$ & $1.53$ & $-71.11$ & $-70.21$ & $-57.34$ & $2.08$  \\
DK & $+1.00$ & $2.11$ & $+6.04$ & $3.55$ & $+23.37$ & $+1{,}440.94$ & $-0.48$ & $0.77$  \\
HR & $+0.09$ & $0.52$ & $+4.04$ & $2.27$ & $-2.73$ & $+0.45$ & $-0.33$ & $0.22$  \\
BE & $+0.52$ & $0.29$ & $+0.91$ & $0.69$ & $-4.36$ & $+14.25$ & $-0.09$ & $0.14$  \\
DE & $-25.10$ & $0.47$ & $-19.81$ & $0.56$ & $-48.88$ & $-30.77$ & $-25.00$ & $0.46$  \\
PT & $-2.84$ & $0.29$ & $-3.49$ & $0.63$ & $-14.98$ & $+8.47$ & $-2.42$ & $0.27$  \\
IT & $-9.76$ & $0.35$ & $-2.54$ & $1.01$ & $-7.79$ & $+0.73$ & $-10.06$ & $0.32$  \\
\midrule
Mean & \multicolumn{2}{r}{$-11.02$\hphantom{\,$\pm$\,0.19}} & \multicolumn{2}{r}{$-8.23$\hphantom{\,$\pm$\,1.53}} & $-18.07$ & $+194.84$ & \multicolumn{2}{r}{$-13.68$\hphantom{\,$\pm$\,2.08}}  \\
Worst & \multicolumn{2}{r}{$+3.45$\hphantom{\,$\pm$\,0.19}} & \multicolumn{2}{r}{$+10.92$\hphantom{\,$\pm$\,1.53}} & $+23.37$ & $+1{,}440.94$ & \multicolumn{2}{r}{$+0.37$\hphantom{\,$\pm$\,2.08}}  \\
Improved & \multicolumn{2}{r}{4/7\hphantom{\,$\pm$\,0.19}} & \multicolumn{2}{r}{4/7\hphantom{\,$\pm$\,1.53}} & 6/7 & 2/7 & \multicolumn{2}{r}{7/7\hphantom{\,$\pm$\,2.08}}  \\
\bottomrule
\end{tabular}
\end{table}

The error-driven rule recomputes $w_{k} \propto 1/\bar\ell_{k}$ on the simplex at every origin, with $\bar\ell_k$ the mean squared error of expert $k$ over the matured origins in the trailing window. Per-origin cross terms are not stored, so the entry reported is $\bar{w}^\top M \bar{w}$ at the time-averaged weight; the realized weights move very little, so this is an approximation in that rule's favor.

\begin{table}[!ht]
\caption{The two-expert gate on the 14 foundation-model pairs. MSE change versus base model in percent, mean $\pm$ standard deviation over five runs, all origins, on the basis of Table~3 of the main text; the standard deviation carries one more decimal than the mean.}
\label{tab:fm-k2}
\centering
\footnotesize
\begin{tabular}{lrr}
\toprule
Dataset & Chronos-Bolt & TimesFM  \\
\midrule
ETTh1 & $-0.67 \pm 0.269$ & $+0.32 \pm 0.117$  \\
ETTh2 & $+0.47 \pm 0.099$ & $-0.87 \pm 0.680$  \\
ETTm1 & $-10.77 \pm 0.292$ & $-3.10 \pm 0.188$  \\
ETTm2 & $-11.53 \pm 0.217$ & $-10.46 \pm 0.216$  \\
Weather & $-5.33 \pm 1.499$ & $-3.02 \pm 0.280$  \\
ECL & $-0.77 \pm 0.115$ & $-1.57 \pm 0.045$  \\
Exchange & $+0.00 \pm 0.000$ & $+0.00 \pm 0.000$  \\
\midrule
\textit{Worst} & $+0.47$ & $+0.32$  \\
\bottomrule
\end{tabular}
\end{table}

\begin{table}[!ht]
\caption{Expert receiving the largest mean gate weight, with that weight, five-run mean. On four cells the two correctors sit close enough that which of them receives the largest weight changes with the run; the result files flag those cells.}
\label{tab:whowins}
\centering
\footnotesize
\begin{tabular}{lllll}
\toprule
Dataset & DLinear & PatchTST & Chronos-Bolt & TimesFM  \\
\midrule
ETTh1 & Online (0.621) & Static (0.467) & Online (0.443) & Static (0.485)  \\
ETTh2 & Online (0.689) & Static (0.548) & Static (0.484) & Static (0.419)  \\
ETTm1 & Static (0.623) & Static (0.435) & Online (0.958) & Online (0.470)  \\
ETTm2 & Online (0.986) & Online (0.962) & Online (0.964) & Online (0.998)  \\
Weather & Static (0.898) & Static (0.759) & Static (0.741) & Static (0.629)  \\
ECL & Static (0.956) & Static (0.915) & Static (0.880) & Static (0.841)  \\
Exchange & Static (0.929) & Static (0.982) & Static (0.998) & Static (0.943)  \\
\bottomrule
\end{tabular}
\end{table}

\label{app:errcorr-s}

The error covariance stored for every combination cell checks the conditional analysis of Section~2 of the main text over the 415 combination runs of Section~5.2. The Spearman correlation between $\rho(e_1, e_2)$ and the variance gap is $-0.42$. Half of the cells realize a weight more than 0.2 from the in-sample optimum, at a median cost of 1.2\% of combination variance. On the most extreme cell, with $\rho = 0.968$, the two weights differ by a factor of 30 while the two losses differ by 0.09\%. This is the case in which trimming to the simplex is predicted to do best \citep{radchenko2023similar} and the weight-estimation variance of \citet{claeskens2016puzzle} is largest.

\section{Scale-free measures and statistical base models}

\begin{table}[!ht]
\caption{MASE and RMSSE against a seasonal naive forecaster of period $m$, mean $\pm$ standard deviation over five runs. MASE is the mean of the five per-run values. RMSSE is the root of the run-mean MSE. The standard deviation printed beside it is that of the five per-run roots, whose mean sits below the printed RMSSE by at most 0.013 on Exchange and by less than the last digit elsewhere.}
\label{tab:scalefree}
\centering
\scriptsize
\setlength{\tabcolsep}{3.5pt}
\begin{tabular}{llrr@{\,$\pm$\,}lr@{\,$\pm$\,}lr@{\,$\pm$\,}lr@{\,$\pm$\,}lr@{\,$\pm$\,}lr@{\,$\pm$\,}l}
\toprule
 &  &  & \multicolumn{6}{c}{MASE} & \multicolumn{6}{c}{RMSSE}  \\
\cmidrule(lr){4-9}\cmidrule(lr){10-15}
Base model & Dataset & $m$ & \multicolumn{2}{c}{Frozen} & \multicolumn{2}{c}{Static} & \multicolumn{2}{c}{Gate, $K{=}3$} & \multicolumn{2}{c}{Frozen} & \multicolumn{2}{c}{Static} & \multicolumn{2}{c}{Gate, $K{=}3$}  \\
\midrule
DLinear & ETTh1 & 24 & $0.999$ & $0.013$ & $0.994$ & $0.012$ & $0.977$ & $0.008$ & $0.932$ & $0.006$ & $0.930$ & $0.006$ & $0.922$ & $0.003$  \\
 & ETTh2 & 24 & $1.123$ & $0.006$ & $1.122$ & $0.006$ & $1.089$ & $0.009$ & $0.898$ & $0.006$ & $0.898$ & $0.007$ & $0.895$ & $0.008$  \\
 & ETTm1 & 96 & $0.845$ & $0.006$ & $0.841$ & $0.006$ & $0.835$ & $0.005$ & $0.820$ & $0.003$ & $0.818$ & $0.003$ & $0.819$ & $0.001$  \\
 & ETTm2 & 96 & $0.868$ & $0.025$ & $0.862$ & $0.024$ & $0.804$ & $0.004$ & $0.706$ & $0.013$ & $0.703$ & $0.013$ & $0.679$ & $0.002$  \\
 & Weather & 144 & $0.594$ & $0.012$ & $0.558$ & $0.004$ & $0.560$ & $0.005$ & $0.471$ & $0.002$ & $0.459$ & $0.002$ & $0.460$ & $0.002$  \\
 & ECL & 24 & $0.891$ & $0.008$ & $0.867$ & $0.003$ & $0.868$ & $0.003$ & $0.770$ & $0.003$ & $0.760$ & $0.001$ & $0.760$ & $0.001$  \\
 & Exchange & 1 & $7.379$ & $0.521$ & $6.993$ & $0.238$ & $7.006$ & $0.243$ & $5.945$ & $0.447$ & $5.641$ & $0.256$ & $5.658$ & $0.255$  \\
\midrule
PatchTST & ETTh1 & 24 & $0.966$ & $0.009$ & $0.964$ & $0.009$ & $0.965$ & $0.009$ & $0.917$ & $0.005$ & $0.916$ & $0.005$ & $0.916$ & $0.005$  \\
 & ETTh2 & 24 & $1.105$ & $0.008$ & $1.103$ & $0.008$ & $1.103$ & $0.008$ & $0.902$ & $0.008$ & $0.901$ & $0.008$ & $0.902$ & $0.008$  \\
 & ETTm1 & 96 & $0.820$ & $0.003$ & $0.818$ & $0.002$ & $0.810$ & $0.005$ & $0.802$ & $0.007$ & $0.801$ & $0.008$ & $0.793$ & $0.002$  \\
 & ETTm2 & 96 & $0.838$ & $0.013$ & $0.835$ & $0.013$ & $0.802$ & $0.003$ & $0.700$ & $0.014$ & $0.699$ & $0.014$ & $0.678$ & $0.002$  \\
 & Weather & 144 & $0.505$ & $0.004$ & $0.512$ & $0.003$ & $0.514$ & $0.002$ & $0.440$ & $0.002$ & $0.437$ & $0.001$ & $0.437$ & $0.001$  \\
 & ECL & 24 & $0.838$ & $0.002$ & $0.833$ & $0.002$ & $0.833$ & $0.002$ & $0.742$ & $0.001$ & $0.738$ & $0.001$ & $0.738$ & $0.001$  \\
 & Exchange & 1 & $7.313$ & $0.060$ & $7.159$ & $0.031$ & $7.159$ & $0.031$ & $5.874$ & $0.058$ & $5.653$ & $0.036$ & $5.661$ & $0.036$  \\
\bottomrule
\end{tabular}
\end{table}

\label{app:classical}

Table~\ref{tab:classical} applies the layer to frozen statistical base models on ETTh1, ETTh2, Weather and Electricity: a seasonal naive forecaster at the calendar day and exponential smoothing with additive damped trend and additive seasonality. Both are fitted per channel on the training split, then frozen, and neither varies across runs. The three remaining datasets are omitted because two conventions for the seasonal period coexist in the pipeline and disagree there by enough to change the reference forecaster's error several fold.

Every one of the 120 corrected cells improves, but almost all weight sits on the online expert, which reads the look-back window and is a forecaster in its own right. The warm start converges to the same vertex that the test stream reaches, so this is not an artifact of initialization.

Sorted by the base model's own error, statistical and neural cells lie on one curve. On ETTh1, the base model error spans a factor of 2.86 while the combined error spans 1.15; on Weather, the factors are 2.20 and 1.30. Two of the four statistical cells end above every neural cell's combined error on the same dataset, so the large percentages reflect a worse starting point, not a better end point. Removing the damping makes the smoothing base model 2.2 times worse and moves the combined result by 1.8\%; damping was fixed a priori.

These cells support an association between the base model's error and the gain from correction, not a difference in how the method behaves across families of base models.

\begin{table}[!ht]
\caption{Frozen statistical base models, mean $\pm$ standard deviation over five runs, the standard deviation at one more decimal so that no entry rounds to zero. Change is the MSE change in percent against the base model in the same row; the static corrector, the held-out corrector and the combination are each run alone.}
\label{tab:classical}
\centering
\footnotesize
\begin{tabular}{llrr@{\,$\pm$\,}lr@{\,$\pm$\,}lr@{\,$\pm$\,}l}
\toprule
Dataset & Base model & MSE & \multicolumn{2}{c}{Static} & \multicolumn{2}{c}{Held-out} & \multicolumn{2}{c}{Gate, $K{=}3$}  \\
\midrule
ETTh1 & Seasonal naive & 0.5122 & $-0.74$ & $0.001$ & $-20.10$ & $0.172$ & $\mathbf{-21.40}$ & $0.257$  \\
 & Smoothing & 1.1014 & $-0.96$ & $0.001$ & $-59.33$ & $0.745$ & $\mathbf{-60.04}$ & $0.709$  \\
ETTh2 & Seasonal naive & 0.3905 & $-0.77$ & $0.010$ & $-22.68$ & $2.043$ & $\mathbf{-23.06}$ & $2.086$  \\
 & Smoothing & 0.4059 & $-1.10$ & $0.004$ & $-27.85$ & $0.447$ & $\mathbf{-28.51}$ & $0.388$  \\
Weather & Seasonal naive & 0.3167 & $-9.76$ & $0.063$ & $-35.73$ & $0.462$ & $\mathbf{-37.89}$ & $0.220$  \\
 & Smoothing & 0.3380 & $-13.41$ & $0.066$ & $-42.90$ & $2.488$ & $\mathbf{-44.30}$ & $2.489$  \\
ECL & Seasonal naive & 0.3211 & $-10.03$ & $0.022$ & $-50.59$ & $0.015$ & $\mathbf{-50.73}$ & $0.023$  \\
 & Smoothing & 1.8147 & $-9.98$ & $0.003$ & $-91.23$ & $0.015$ & $\mathbf{-91.29}$ & $0.011$  \\
\bottomrule
\end{tabular}
\end{table}

\section{Sensitivity of the layout constants and the learning rate}

\label{app:layout}

Panel A of Table~\ref{tab:s-layout} halves and doubles, one at a time, each of the three layout constants of Section~4.3 and the trust-region radius on ETTh2 with Chronos-Bolt, the pair with the worst deterioration. The reported configuration occurs four times in the grid and returns $+0.1463$ at each.

\begin{table}[!ht]
\caption{Sensitivity of the gate, mean $\pm$ standard deviation over five runs. A, one-at-a-time perturbation of the layout constants on ETTh2 with Chronos-Bolt, gate $K{=}3$. B, early-stopping tail across the foundation-model pairs, gate $K{=}3$. A and B give the MSE change versus the frozen base model in percent. C, gate learning rate, MSE.}
\label{tab:s-layout}
\centering
\footnotesize
\textit{A}\par\smallskip
\begin{tabular}{llr@{\,$\pm$\,}l}
\toprule
Constant & Grid point & \multicolumn{2}{c}{MSE change (\%)}  \\
\midrule
Trust-region radius $\delta$ & $0.5\times$ & $+0.19$ & $0.076$  \\
 & $1\times$ (reported) & $+0.15$ & $0.081$  \\
 & $2\times$ & $+0.04$ & $0.119$  \\
Warm slice length & $H+100$ & $-0.08$ & $0.026$  \\
 & $H+200$ (reported) & $+0.15$ & $0.081$  \\
 & $H+400$ & $+0.15$ & $0.054$  \\
Early-stopping tail & 5\% & $-0.17$ & $0.032$  \\
 & 10\% (reported) & $+0.15$ & $0.081$  \\
 & 20\% & $-0.15$ & $0.037$  \\
Online update cadence & Every 32 matured origins & $+0.08$ & $0.103$  \\
 & Every 64 (reported) & $+0.15$ & $0.081$  \\
 & Every 128 & $+0.17$ & $0.078$  \\
\bottomrule
\end{tabular}

\medskip
\textit{B}\par\smallskip
\begin{tabular}{llr@{\,$\pm$\,}lr@{\,$\pm$\,}lr@{\,$\pm$\,}l}
\toprule
Dataset & Base model & \multicolumn{2}{c}{5\%} & \multicolumn{2}{c}{10\% (reported)} & \multicolumn{2}{c}{20\%}  \\
\midrule
ETTh1 & Chronos-Bolt & $-2.08$ & $0.387$ & $-0.80$ & $0.265$ & $-0.32$ & $0.079$  \\
ETTh2 & Chronos-Bolt & $-0.17$ & $0.032$ & $+0.15$ & $0.081$ & $-0.15$ & $0.037$  \\
ETTm1 & Chronos-Bolt & $-10.16$ & $0.225$ & $-10.55$ & $0.288$ & $-8.77$ & $0.441$  \\
ETTm2 & Chronos-Bolt & $-11.75$ & $0.097$ & $-11.53$ & $0.217$ & $-10.63$ & $0.802$  \\
Weather & Chronos-Bolt & $-8.18$ & $0.163$ & $-8.61$ & $0.464$ & $-8.51$ & $0.318$  \\
ECL & Chronos-Bolt & $-0.78$ & $0.066$ & $-0.82$ & $0.068$ & $-0.89$ & $0.061$  \\
Exchange & Chronos-Bolt & $-7.95$ & $0.190$ & $-8.09$ & $0.213$ & $-4.29$ & $0.692$  \\
ETTh1 & TimesFM & $-2.46$ & $0.390$ & $+0.04$ & $0.062$ & $-0.07$ & $0.066$  \\
ETTh2 & TimesFM & $-1.23$ & $0.714$ & $-0.86$ & $0.628$ & $-0.35$ & $0.341$  \\
ETTm1 & TimesFM & $-1.55$ & $0.589$ & $-3.15$ & $0.546$ & $-0.13$ & $0.248$  \\
ETTm2 & TimesFM & $-10.70$ & $0.088$ & $-10.46$ & $0.217$ & $-9.94$ & $0.302$  \\
Weather & TimesFM & $-5.01$ & $0.067$ & $-5.54$ & $0.054$ & $-5.31$ & $0.125$  \\
ECL & TimesFM & $-1.32$ & $0.231$ & $-1.70$ & $0.079$ & $-1.75$ & $0.045$  \\
Exchange & TimesFM & $-3.98$ & $0.443$ & $-4.43$ & $0.301$ & $+0.56$ & $3.235$  \\
\midrule
\textit{Mean} &  & \multicolumn{2}{r}{$-4.81$} & \multicolumn{2}{r}{$-4.74$} & \multicolumn{2}{r}{$-3.61$}  \\
\textit{Worst} &  & \multicolumn{2}{r}{$-0.17$} & \multicolumn{2}{r}{$+0.15$} & \multicolumn{2}{r}{$+0.56$}  \\
\bottomrule
\end{tabular}

\medskip
\textit{C}\par\smallskip
\begin{tabular}{lr@{\,$\pm$\,}lr@{\,$\pm$\,}lr@{\,$\pm$\,}l}
\toprule
Cell & \multicolumn{2}{c}{$\eta{=}0.05$} & \multicolumn{2}{c}{$\eta{=}0.1$} & \multicolumn{2}{c}{$\eta{=}0.3$}  \\
\midrule
ETTh2 $\times$ DLinear & $0.28742$ & $0.004919$ & $0.28930$ & $0.004940$ & $0.29070$ & $0.004968$  \\
Exchange $\times$ DLinear & $0.09143$ & $0.008558$ & $0.09095$ & $0.008214$ & $0.09066$ & $0.007925$  \\
ETTm2 $\times$ Chronos-Bolt & $0.16201$ & $0.000398$ & $0.16197$ & $0.000398$ & $0.16199$ & $0.000393$  \\
\bottomrule
\end{tabular}
\end{table}

Two grid points exceed it: $\delta$ at $0.5\times$ ($+0.1924$) and the update interval at 128 ($+0.1677$).

Panel B of Table~\ref{tab:s-layout} varies the early-stopping tail across the 14 foundation-model pairs: the 5\% tail improves every pair, the reported 10\% tail 12, and the 20\% tail 13. At 5\%, no pair that the reported setting improves deteriorates. At 20\%, exactly one does: Exchange with TimesFM, the pair with the smallest fit region.

Panel C of Table~\ref{tab:s-layout} gives the gate learning rate sweep behind Section~5.5 of the main text.

Larger $\eta$ sharpens the allocation toward a vertex without reaching it, the convexity limitation of Section~7.3 of the main text; $\eta = 0.1$ is kept everywhere.

\section{Interval scores: native quantiles and load pinball}

\begin{figure}[!ht]
\centering
\includegraphics[width=\textwidth]{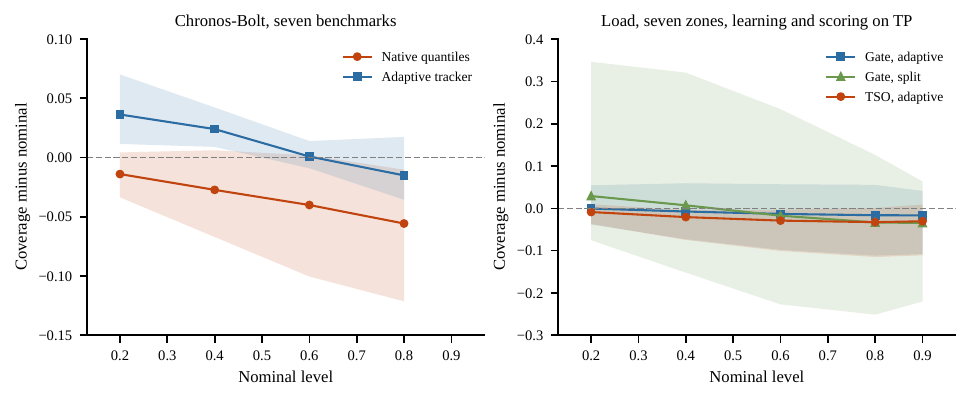}
\caption{Empirical coverage minus nominal level, by dataset and zone. Left, the native quantile head and the adaptive tracker on frozen Chronos-Bolt; right, the three interval methods of Table~7 of the main text on the load data. Lines are means over datasets or zones, bands the range across those datasets or zones, and the dashed line perfect calibration.}
\label{fig:reliability}
\end{figure}

Table~\ref{tab:intervals} and Fig.~\ref{fig:reliability} report the interval comparison on frozen Chronos-Bolt of Section~5.5 of the main text, the right panel extending the load comparison of Section~6.4 to every level. The native head under-covers by a gap that widens with the level, from 0.014 at 0.2 to 0.056 at 0.8, while the tracker stays within 0.04 of nominal on average. Table~\ref{tab:opsd-pinball} scores the load interval methods of Table~7 by mean pinball loss.

\begin{table}[!ht]
\caption{Intervals at a matched 80\% level on frozen Chronos-Bolt, with the mean pinball loss over the nine deciles. Coverage closer to 0.80 and lower pinball are better. Neither method carries a random component on a frozen zero-shot forecaster, so the five runs coincide and no dispersion is reported.}
\label{tab:intervals}
\centering
\footnotesize
\begin{tabular}{lrrrrrr}
\toprule
 & \multicolumn{3}{c}{Native quantiles} & \multicolumn{3}{c}{Adaptive tracker}  \\
\cmidrule(lr){2-4}\cmidrule(lr){5-7}
Dataset & Cov & Width & Pinball & Cov & Width & Pinball  \\
\midrule
ETTh1 & 0.759 & 1.066 & 0.152 & 0.817 & 1.226 & 0.159  \\
ETTh2 & 0.739 & 0.833 & 0.132 & 0.764 & 0.938 & 0.137  \\
ETTm1 & 0.746 & 0.902 & 0.137 & 0.797 & 1.021 & 0.141  \\
ETTm2 & 0.757 & 0.669 & 0.100 & 0.775 & 0.808 & 0.105  \\
Weather & 0.741 & 0.576 & 0.088 & 0.776 & 0.693 & 0.090  \\
ECL & 0.790 & 0.622 & 0.085 & 0.790 & 0.646 & 0.088  \\
Exchange & 0.679 & 0.545 & 0.092 & 0.775 & 0.730 & 0.091  \\
\midrule
Mean $|$cov${-}0.80|$ & \multicolumn{3}{c}{0.056} & \multicolumn{3}{c}{$\mathbf{0.020}$}  \\
Mean pinball & \multicolumn{3}{c}{$\mathbf{0.112}$} & \multicolumn{3}{c}{0.116}  \\
\bottomrule
\end{tabular}
\end{table}

\begin{table}[!ht]
\caption{Mean pinball loss over the nine deciles, same three methods and basis as Table~7 of the main text. Lower is better. Gate columns print mean $\pm$ standard deviation over five runs; the TSO forecast and its split-calibrated width carry no run variation.}
\label{tab:opsd-pinball}
\centering
\footnotesize
\begin{tabular}{lr@{\,$\pm$\,}lr@{\,$\pm$\,}lr}
\toprule
Zone & \multicolumn{2}{c}{Gate, adaptive} & \multicolumn{2}{c}{Gate, split} & TSO, adaptive  \\
\midrule
HU & $\mathbf{0.06316}$ & $0.001486$ & 0.06981 & $0.001036$ & 0.09910  \\
PT & $\mathbf{0.09742}$ & $0.000127$ & 0.10313 & $0.000152$ & 0.09871  \\
HR & $\mathbf{0.04708}$ & $0.000068$ & 0.04711 & $0.000066$ & 0.04716  \\
BE & $\mathbf{0.05774}$ & $0.000035$ & 0.05794 & $0.000042$ & 0.05777  \\
DK & $\mathbf{0.02282}$ & $0.000108$ & 0.02308 & $0.000111$ & 0.02286  \\
DE & $\mathbf{0.07294}$ & $0.000220$ & 0.07361 & $0.000247$ & 0.08553  \\
IT & $\mathbf{0.03089}$ & $0.000057$ & 0.03095 & $0.000052$ & 0.03287  \\
\midrule
Mean & \multicolumn{2}{r}{$\mathbf{0.05601}$} & \multicolumn{2}{r}{0.05795} & 0.06343  \\
\bottomrule
\end{tabular}
\end{table}

\section{Zone selection and the load layout}

\label{app:opsd-s}

Of the 36 bidding zones with a day-ahead forecast and both actual-load versions, seven pass three screens: no series with 1\% or more missing values; no gap longer than three hours in either outcome series, shorter gaps linearly interpolated; and no TSO forecast hour above three times the concurrent load. The settled outcome is among the binding series for 16 of the 29 rejected zones and the only binding series for eight, the publication delay appearing as missing data. Table~\ref{tab:app-zones} lists every zone with its outcome.

Forecast holes longer than three hours cost origins rather than a zone: 129 hours in Germany, 126 in Belgium, 21 each in Croatia and Italy. The common grid is 2015-01-05 00:00 to 2019-01-31 21:00 UTC, 35{,}710 hours, complete in both outcome versions.

\begin{table}[!ht]
\caption{Accepted zones and the cause of each rejection. Missing fractions are over the common grid; defect hours are forecast hours above three times load.}
\label{tab:app-zones}
\centering
\footnotesize
\begin{tabular}{llrrr}
\toprule
Zone & Verdict & Miss prov. & Miss settled & Defect h  \\
\midrule
DE, HU, PT, HR, DK, IT, BE & Accept & $\le 0.0001$ & $\le 0.0001$ & 0  \\
\midrule
NL & Defect & 0.0000 & 0.0001 & 96  \\
CH & Defect & 0.0000 & 0.0000 & 20  \\
ES, SK, FR, RO, FI, LV, CZ, ME & Gap & $\le 0.0062$ & $\le 0.0000$ & $\le 10$  \\
AT, SI, PL, GR, NO, SE & Missing settled & $\le 0.0041$ & $\ge 0.0209$ &  \\
GB\_GBN, DK\_1 & Missing settled & 0.0000 & $\ge 0.7574$ &  \\
LT, BG & Missing forecast & $\le 0.0036$ & $\le 0.0001$ &  \\
EE, IE, LU, RS, MK, GB\_UKM, CY, GB\_NIR, UA\_west & Missing several & Up to 0.674 & Up to 0.757 &  \\
\bottomrule
\end{tabular}
\end{table}

\begin{table}[!ht]
\caption{Held-out layout. A, load streams in daily origins: the warm slice is $1+D+200$ origins and the early-stopping tail 30 origins, the fit region the remainder; Belgium's held-out split is two origins shorter than the other six zones'. At $D{=}90$ the slice alone would take 291 origins, so that configuration is refused (Section~7.2.2 of the main text). B, benchmarks by horizon: held-out split and the fit region remaining after the warm slice and the tail.}
\label{tab:load-layout}
\centering
\footnotesize
\textit{A}\par\smallskip
\begin{tabular}{lrrrr}
\toprule
 & Held-out & Warm slice & Tail & Fit region  \\
\midrule
$D{=}0$, six zones & 297 & 201 & 30 & 66  \\
$D{=}0$, Belgium & 295 & 201 & 30 & 64  \\
$D{=}30$, six zones & 297 & 231 & 30 & 36  \\
$D{=}30$, Belgium & 295 & 231 & 30 & 34  \\
\bottomrule
\end{tabular}

\medskip
\textit{B}\par\smallskip
\begin{tabular}{lrrrrrr}
\toprule
 & \multicolumn{2}{c}{$H{=}96$} & \multicolumn{2}{c}{$H{=}192$} & \multicolumn{2}{c}{$H{=}336$}  \\
Dataset & Split & Fit & Split & Fit & Split & Fit  \\
\midrule
Exchange & 665 & 303 & 569 & 120 & 425 & Refused  \\
ECL & 2537 & 1987 & 2441 & 1805 & 2297 & 1531  \\
ETTh1, ETTh2 & 2785 & 2211 & 2689 & 2028 & 2545 & 1755  \\
Weather & 5175 & 4361 & 5079 & 4179 & 4935 & 3905  \\
ETTm1, ETTm2 & 11425 & 9987 & 11329 & 9804 & 11185 & 9531  \\
\bottomrule
\end{tabular}
\end{table}

The forecast-defect screen is post-hoc for the Netherlands and pre-specified for the other 30 zones. The forecast column for the Netherlands contains 96 hours peaking at 509{,}191 MW against a load of 13{,}548 MW. Section~6 of the main text reports that zone as a sensitivity, and adding it does not change any figure reported for the seven zones. The ratio of the forecast's standard deviation to the load's is 0.94 to 1.03 for every accepted zone and 8.94 for the Netherlands, so no threshold between those values changes the selection. Table~\ref{tab:app-labels} reports the two outcome versions and the forecast quality of each accepted zone.

\begin{table}[!ht]
\caption{Outcome versions and forecast quality per accepted zone, whole window. Revision is the mean absolute relative difference between the two outcome versions, the column Table~8 of the main text also carries; MAPE vs prov. is the forecast's own mean absolute percentage error against that version and zMSE is the mean squared error on per-channel z-scaled series. The entries describe the data and not a fitted model, so they carry no run dispersion.}
\label{tab:app-labels}
\centering
\footnotesize
\begin{tabular}{lrrrrr}
\toprule
Zone & Mean settled/prov. & Revision (\%) & MAPE vs prov. & zMSE vs prov. & zMSE vs settled  \\
\midrule
DE & 1.0573 & 5.930 & 3.104 & 0.0508 & 0.2023  \\
HU & 0.9942 & 0.592 & 3.846 & 0.0870 & 0.0722  \\
PT & 1.0003 & 2.225 & 2.773 & 0.0497 & 0.0990  \\
HR & 1.0235 & 2.598 & 2.088 & 0.0226 & 0.0795  \\
DK & 1.0352 & 3.870 & 1.017 & 0.0089 & 0.0608  \\
IT & 1.0940 & 9.401 & 1.973 & 0.0135 & 0.1576  \\
BE & 0.9769 & 3.183 & 2.372 & 0.0492 & 0.1226  \\
\bottomrule
\end{tabular}
\end{table}

Italy is the extreme case. Its TSO forecast is the second most accurate against the provisional outcome but the second least accurate against the settled one. The settled load exceeds the provisional load almost everywhere, with a first-percentile ratio of 1.044. Every zone except Hungary scores materially worse against the settled outcome.

\section{Short series}

\label{app:short}

The weekly influenza-like-illness series has 966 rows. Its held-out split supplies $194 - H + 1$ origins at a 6:2:2 split and $97 - H + 1$ at 7:1:2, below the 201 that a warm slice of $H+200$ needs even at $H = 0$. At the most generous split and the shortest horizon, the series would need 1{,}358 rows, 1.41 times as many as it has. The gate and the plain held-out corrector consume the same layout, so all 30 cells are skipped with their origin counts recorded.

Table~\ref{tab:app-short} compares a shortened warm start against a uniform initialization on an identical layout and fitted corrector, so that only the gate's starting weights differ.

\begin{table}[!ht]
\caption{Short-series warm start, five runs; the three MSE columns print mean $\pm$ standard deviation. The last column is the uniform column minus the warm column, negative meaning the warm start is worse.}
\label{tab:app-short}
\centering
\footnotesize
\begin{tabular}{lrrr@{\,$\pm$\,}lr@{\,$\pm$\,}lr@{\,$\pm$\,}l}
\toprule
Split, $L$, $H$ & Slice & Updates & \multicolumn{2}{c}{Warm} & \multicolumn{2}{c}{Uniform} & \multicolumn{2}{c}{Uniform $-$ warm}  \\
\midrule
6:2:2, 36, 24 & 77 & 53 & $-6.47$ & $1.845$ & $-6.85$ & $2.044$ & $-0.38$ & $0.214$  \\
6:2:2, 36, 36 & 71 & 35 & $-14.32$ & $0.700$ & $-15.82$ & $0.696$ & $-1.50$ & $0.140$  \\
6:2:2, 36, 48 & 66 & 18 & $-18.98$ & $0.965$ & $-20.73$ & $0.980$ & $-1.75$ & $0.241$  \\
6:2:2, 104, 24 & 77 & 53 & $-8.28$ & $0.943$ & $-9.07$ & $1.022$ & $-0.79$ & $0.183$  \\
7:1:2, 36, 24 & 33 & 9 & $-1.00$ & $0.775$ & $-1.00$ & $0.764$ & $-0.00$ & $0.015$  \\
\bottomrule
\end{tabular}
\end{table}

The warm-started allocation, $(0.073, 0.072, 0.855)$, is indistinguishable from the uniformly started one, $(0.074, 0.072, 0.854)$: a slice of 66 to 77 origins evaluates the online corrector before it has adapted. The largest slice the data allows is worse, $(0.47, 0.45, 0.08)$ after 129 replay updates against a corrector fitted on a single origin. On run means no variant deteriorates; the largest per-run deterioration across the 150 variant cells is 0.24\%.

In all 50 matched comparisons, the plain held-out corrector beats the gate on the identical layout by 1.01 to 5.46 points. With 98 to 146 matured test origins, the gate lacks the feedback to learn the allocation, so on streams this short the combination is optional rather than assumed.

The same limit binds on the load data at the ninety-day publication delay (Section~6 of the main text); every corrector method there records a skipped cell with its origin count.

\section{Longer horizons}

\label{app:horizon}

Horizons 192 and 336 repeat the horizon-96 protocol on the two trained base models. Four of the 14 pairs of Table~2 of the main text have no rows at either horizon, so 10 pairs carry horizon 192 and nine carry 336. These results support the layout condition of Section~7.2 of the main text, not a claim about horizon generalization.

Table~\ref{tab:app-horizon} gives the held-out corrector and the combination against their frozen base model. Each entry carries the standard deviation of the five per-run changes, and nine of the 19 pairs have a gate change smaller than that standard deviation and are marked.

\begin{table}[!ht]
\caption{Combination against frozen base model at horizons 192 and 336, MSE change in percent, five-run mean with the standard deviation of the five per-run changes after $\pm$. Negative is better. A dagger marks a gate change smaller in absolute value than that standard deviation.}
\label{tab:app-horizon}
\centering
\footnotesize
\begin{tabular}{llrrl}
\toprule
Dataset & Base model & Held-out & Gate, $K{=}3$ &  \\
\midrule
\multicolumn{5}{l}{\textit{Horizon 192}} \\
Exchange & DLinear & $+90.12 \pm 18.635$ & $+23.03 \pm 14.989$ &  \\
ETTh2 & DLinear & $+3.24 \pm 2.833$ & $+1.01 \pm 1.616$ & $\dagger$  \\
ETTh2 & PatchTST & $+2.23 \pm 2.746$ & $+0.11 \pm 1.543$ & $\dagger$  \\
ETTh1 & PatchTST & $+9.60 \pm 2.087$ & $+0.02 \pm 0.281$ & $\dagger$  \\
ETTm1 & DLinear & $+1.95 \pm 0.984$ & $-0.50 \pm 0.758$ & $\dagger$  \\
ETTh1 & DLinear & $+5.37 \pm 4.934$ & $-2.17 \pm 2.998$ & $\dagger$  \\
ETTm1 & PatchTST & $-1.90 \pm 1.352$ & $-3.90 \pm 1.469$ &  \\
Weather & DLinear & $+6.65 \pm 4.341$ & $-5.27 \pm 0.988$ &  \\
ETTm2 & DLinear & $-6.03 \pm 2.433$ & $-5.71 \pm 2.408$ &  \\
ETTm2 & PatchTST & $-6.04 \pm 2.313$ & $-5.73 \pm 2.343$ &  \\
\midrule
\multicolumn{5}{l}{\textit{Horizon 336}} \\
ETTh1 & DLinear & $+18.18 \pm 1.739$ & $-0.17 \pm 0.073$ &  \\
ETTh2 & PatchTST & $+2.83 \pm 1.696$ & $-0.26 \pm 0.632$ & $\dagger$  \\
ETTm1 & DLinear & $+2.61 \pm 2.054$ & $-1.39 \pm 1.987$ & $\dagger$  \\
ETTh1 & PatchTST & $+16.91 \pm 6.837$ & $-1.46 \pm 1.909$ & $\dagger$  \\
ETTm2 & PatchTST & $-3.58 \pm 3.648$ & $-2.72 \pm 3.569$ & $\dagger$  \\
ETTm1 & PatchTST & $+1.16 \pm 1.630$ & $-2.95 \pm 0.875$ &  \\
Weather & DLinear & $+13.27 \pm 4.720$ & $-3.48 \pm 1.642$ &  \\
ETTh2 & DLinear & $-3.25 \pm 5.597$ & $-3.90 \pm 3.247$ &  \\
ETTm2 & DLinear & $-4.18 \pm 2.558$ & $-4.60 \pm 2.163$ &  \\
\bottomrule
\end{tabular}
\end{table}

At horizon 336, no pair deteriorates; Exchange is absent because the layout guard refuses that pair. At horizon 192, four pairs deteriorate, three of them by less than their own run spread. The same criterion marks three pairs at horizon 96, so gains within run noise are not specific to the longer horizons.

Panel B of Table~\ref{tab:load-layout} gives the held-out layout behind those cells: the warm slice grows with the horizon while the held-out split shrinks, and the fit region absorbs the difference.

\section{Expert speed and the fourth expert}

Panel A of Table~\ref{tab:quasistatic} gives the post-hoc learning-rate diagnosis behind Section~7.1 of the main text, and panel B the intercept correction admitted as a fourth expert.

\begin{table}[!ht]
\caption{Expert speed and the fourth expert, MSE change versus the frozen base model in percent. A, post-hoc diagnosis on ETTh2 with DLinear: slowing the added expert improves that expert on its own and restores the gate's downside control at the same time. B, the intercept correction as a fourth expert, five-run means: worst pair over the 28 benchmark pairs, worst zone and mean over the seven zones; bases of Tables~2 and~6 of the main text. A dash marks a configuration that was not run.}
\label{tab:quasistatic}
\centering
\footnotesize
\textit{A}\par\smallskip
\begin{tabular}{lrr}
\toprule
Expert learning rate & Expert alone & Four-expert gate  \\
\midrule
0.005 (published default) & $-0.11$ & $+1.22$  \\
0.0005 & $-0.83$ & $-0.40$  \\
0.00005 & $-0.09$ & $-0.63$  \\
\bottomrule
\end{tabular}

\medskip
\textit{B}\par\smallskip
\begin{tabular}{lrrr}
\toprule
 & Gate, $K{=}3$ & Intercept alone & Gate, $K{=}4$  \\
\midrule
Benchmarks, worst pair & $+0.15$ & $+33.08$ & $+0.47$  \\
Load L1, worst zone & $-0.09$ & $+23.37$ & $+5.00$  \\
Load L1, mean & $-13.68$ & $-18.07$ & $\mathbf{-20.89}$  \\
Load L3 $D{=}30$, worst zone & $-6.50$ & -- & $+16.00$  \\
\bottomrule
\end{tabular}
\end{table}

\section{Alternative aggregation rules over the same experts}
\label{sec:aggrules}

Table~\ref{tab:aggrules} replays three alternative weight updates over the expert forecasts of the main experiments: fixed-share \citep{herbster1998} at the $\eta$ of the main experiments with switching rates $\alpha = 0.01$ and $\alpha = 1/T_m$, where $T_m$ is the stream's number of matured test origins, and Bernstein online aggregation \citep{wintenberger2017boa} with its self-tuned rates. Each rule is warm-started by replaying its own update on the same warm slice, and both $\alpha$ values are reported with no selection between them. All rules, the Hedge gate included, are priced at their time-averaged weight, $\bar{w}^{\top} M \bar{w}$ over the stored per-origin error moments. That price is exact for a constant weight and bounds a moving one from above, so rules that move their weights more are penalized more, and these entries cannot be set beside the streaming figures of Tables~2 and~5. Against the Hedge gate on the same basis, the gap is $-0.05$ percentage points for fixed-share at $1/T_m$ and $+0.15$ for Bernstein aggregation on the benchmark pairs, both inside the 0.52-point mean pricing gap; on the load cells the Bernstein gap is $+0.01$ points under L1 and $+0.98$ under L3 at $D=30$, with a worst zone of $+3.29$. The result files also carry a fixed-rate Bernstein variant at $\eta = 0.1$.

\begin{table}[!ht]
\caption{Aggregation rules replayed over the same experts, MSE change versus the frozen base model in percent at the time-averaged weight. A, the 36 benchmark pairs at horizon 96, the 28 main-grid pairs and the 8 classical-base-model pairs of Table~\ref{tab:classical}; B, the 21 load zone-protocol cells. Pair and cell entries are five-run means; vs gate is the mean gap to the replayed Hedge gate in percentage points; better counts pairs or cells below the gate.}
\label{tab:aggrules}
\centering
\begin{tabular}{lrrrr}
\toprule
Rule & Mean & vs gate & Better & Worst \\
\midrule
\multicolumn{5}{l}{A. Benchmarks, 36 pairs} \\
Hedge (the gate) & $-13.44$ & -- & -- & $-0.17$ \\
Fixed-share, $\alpha = 1/T_m$ & $-13.49$ & $-0.05$ & 19/36 & $-0.24$ \\
Fixed-share, $\alpha = 0.01$ & $-12.02$ & $+1.41$ & 6/36 & $-0.16$ \\
Bernstein aggregation & $-13.29$ & $+0.15$ & 17/36 & $-0.00$ \\
\midrule
\multicolumn{5}{l}{B. Load, 21 zone-protocol cells} \\
Hedge (the gate) & $-24.15$ & -- & -- & $-0.08$ \\
Fixed-share, $\alpha = 1/T_m$ & $-24.42$ & $-0.27$ & 12/21 & $-0.27$ \\
Fixed-share, $\alpha = 0.01$ & $-23.79$ & $+0.36$ & 8/21 & $-0.02$ \\
Bernstein aggregation & $-23.68$ & $+0.48$ & 7/21 & $-0.00$ \\
\bottomrule
\end{tabular}
\end{table}

\section{Rank test with the dataset as the block}
\label{sec:friedman-dataset}

Table~\ref{tab:friedman-dataset} repeats the Friedman and Nemenyi analysis of Table~5 with the dataset as the block. Within each dataset, a method's MSE change is the mean over the base models in that block set, ranked across methods; there are seven blocks, so the critical distance widens. The DLinear-only set already used seven blocks and is unchanged.

\begin{table}[!ht]
\caption{Friedman test with seven dataset blocks: rank of the three-expert gate, Friedman $p$, Nemenyi critical distance (CD) at the 5\% level, and the separations that remain, beside the pair-block values of Table~5.}
\label{tab:friedman-dataset}
\centering
\begin{tabular}{lrrrrl}
\toprule
Block set & Pair-block $p$ & Dataset-block $p$ & Gate rank & CD & Separations \\
\midrule
All base models & $3.1 \times 10^{-8}$ & $0.008$ & 1.571 & 1.773 & gate vs base model \\
Trained base models & $0.001$ & $0.038$ & 1.857 & 2.306 & gate vs base model \\
DLinear only & $0.246$ & $0.246$ & 3.143 & 3.405 & none \\
\bottomrule
\end{tabular}
\end{table}

\bibliographystyle{elsarticle-harv}
\bibliography{refs}